\documentclass[10pt,twocolumn,letterpaper]{article}

\usepackage[pagenumbers]{extend} 

\usepackage{pgfplots}
\pgfplotsset{compat=1.18}
\usepgfplotslibrary{groupplots}
\usetikzlibrary{positioning, arrows.meta, fit, calc, backgrounds}
\usepackage{xspace}
\newcommand{\rdb}{RealDocBench\xspace}

\newcommand{\lmark}[2]{\tikz[baseline=-0.5ex]{\draw[#1] plot[mark=#2,
  mark size=2.8pt, mark options={fill=#1, draw=#1}] coordinates {(0,0)};}}
\newcommand{\lmarkbig}[2]{\tikz[baseline=-0.5ex]{\draw[#1, very thick]
  plot[mark=#2, mark size=4pt] coordinates {(0,0)};}}

\newcommand{\cmark}{\textcolor{green!55!black}{\checkmark}}
\newcommand{\xmark}{\textcolor{red}{\ensuremath{\times}}}

\definecolor{cvprblue}{rgb}{0.21,0.49,0.74}
\usepackage[pagebackref,breaklinks,colorlinks,allcolors=cvprblue]{hyperref}

\def\paperID{*****} 
\def\confName{CVPR}
\def\confYear{2026}

\title{RealDocBench: A Benchmark for Field-Level QA and Layout
Understanding on Real-World Regulated Documents}

\author{
Ameya Joshi, Joon Kim, Gus Eggert, Joseph Bajor, Cindy Hao \\
Jing Reyhan, Kushal Byatnal, Eli Badgio \\
Extend AI \\
{\tt\small{\{ameya, joon, gus, joe, cindy, jing, kushal, eli\}@extend.ai}}
}   

\begin{document}
\maketitle

\begin{abstract}
Document parsing systems are increasingly deployed in high-stakes, regulated
workflows such as mortgage underwriting, financial reporting, supply-chain logistics,
and clinical records. Yet most public benchmarks evaluate parsers on clean
academic layouts or synthetic prose, and report a single OCR- or
markdown-level similarity score. Such documents and metrics correlate poorly with what
downstream agents actually need: the correct \emph{value} for a specific
\emph{field} on a messy real-world page. We introduce \rdb, a two-track
benchmark built from real regulated documents. The
\textbf{QA track} contains 1{,}356 field-level questions over 581
documents spanning four domains, where each question is paired with a typed
\texttt{gold\_dict} of key$\rightarrow$value answers and parsers are scored on
both per-field and strict per-question accuracy. The \textbf{layout track}
contains 1{,}500 human-verified page images annotated with COCO-style bounding
boxes under a nine-class public taxonomy, scored with a Hungarian matcher that
includes adjacency-aware split/merge recovery. We evaluate eighteen systems,
spanning commercial parsing APIs, general-purpose VLMs, and open-source OCR
models, under a uniform extraction-and-scoring protocol, and report accuracy
alongside per-page cost and cache-busted latency. \rdb exposes a wide
performance spread that single-number benchmarks hide, a persistently hard medical sub-domain, and
sharp cost/latency trade-offs across operating points. We release the datasets, parser
adapters, and evaluation harness to support reproducible, field-level
comparison of document parsing systems.
\end{abstract}

\section{Introduction}
\label{sec:intro}

\begin{figure*}[t]
  \centering
  \resizebox{\textwidth}{!}{%
  \begin{tikzpicture}[
    font=\footnotesize,
    >={Latex[length=2.2mm]},
    docnode/.style={draw=black!55, inner sep=0, line width=0.6pt},
    card/.style={draw, rounded corners, align=center, inner sep=5pt,
      minimum height=1.6cm},
    qa/.style={card, fill=cvprblue!8, draw=cvprblue!60},
    lay/.style={card, fill=orange!8, draw=orange!65},
    score/.style={draw=green!45!black, rounded corners, align=center,
      inner sep=5pt, fill=green!12, minimum height=1.6cm, font=\small},
    flow/.style={-{Latex[length=2.4mm]}, line width=1pt},
    lane/.style={rounded corners=6pt, line width=0.8pt},
    lanelbl/.style={font=\small\bfseries},
  ]
    \node[docnode] (doc)
      {\includegraphics[width=2.2cm]{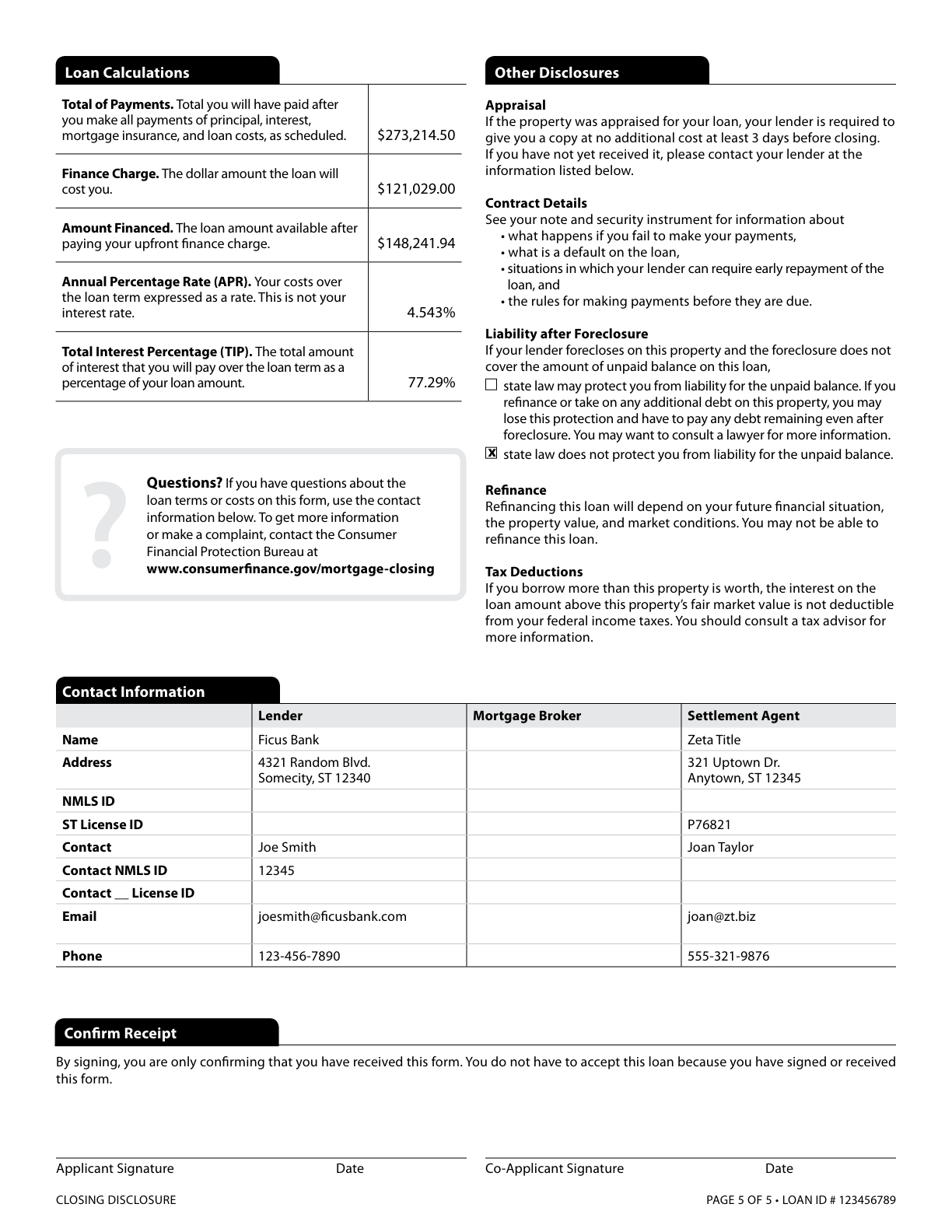}};
    \node[below=2pt of doc, align=center, font=\scriptsize, text width=2.5cm]
      {Real regulated documents\\(4 domains, 581 docs)};

    \node[qa, above right=0.35cm and 1.5cm of doc, anchor=west,
      text width=3.0cm] (qbox)
      {\textbf{Field-level question}\\[1pt]
       \textit{``Lender's email and Settlement Agent ST license ID?''}\\[2pt]
       {\scriptsize typed \texttt{gold\_dict}:}\\
       {\scriptsize\ttfamily \{lender\_email:\,..., st\_license\_id:\,...\}}};
    \node[qa, right=0.7cm of qbox, text width=2.5cm] (pbox)
      {\textbf{16 systems}\\parse $\rightarrow$ markdown\\[2pt]
       {\scriptsize commercial APIs\\ VLMs · open-source OCR}};
    \node[qa, right=0.7cm of pbox, text width=2.5cm] (ebox)
      {\textbf{Extraction LLM}\\$+$ typed scorer\\[2pt]
       {\scriptsize type-tolerant\\ per-field \& per-question}};
    \node[score, right=0.7cm of ebox, text width=1.9cm] (sbox)
      {\textbf{96.0}\\{\scriptsize per-field}\\[2pt]
       \textbf{90.9}\\{\scriptsize per-question}};

    \node[lay, below right=0.35cm and 1.5cm of doc, anchor=west,
      text width=2.6cm] (lbox)
      {\textbf{Page $\rightarrow$ regions}\\[3pt]
       \tikz[baseline, x=1cm, y=1cm, line width=0.5pt]{
         \draw[black!30, fill=white] (0,0) rectangle (1.7,1.45);
         \draw[gray!70, fill=gray!18]        (0.08,1.30) rectangle (1.62,1.40);
         \draw[orange!75, fill=orange!18]    (0.08,1.08) rectangle (1.10,1.24);
         \draw[cvprblue!70, fill=cvprblue!15](0.08,0.72) rectangle (1.62,1.02);
         \draw[green!50!black, fill=green!18](0.08,0.50) rectangle (1.62,0.66);
         \draw[violet!70, fill=violet!15]    (0.08,0.10) rectangle (1.62,0.44);
       }};
    \node[lay, right=0.7cm of lbox, text width=2.5cm] (tbox)
      {\textbf{9-class\\taxonomy}\\[2pt]
       {\scriptsize text · heading · section\\ header · footer · page\,no.\\
        figure · table · key--value}};
    \node[lay, right=0.7cm of tbox, text width=2.5cm] (mbox)
      {\textbf{Adjacency-aware\\ matcher}\\[2pt]
       {\scriptsize Hungarian, IoU\,$\ge$\,0.5\\ split/merge recovery}};
    \node[score, right=0.7cm of mbox, text width=1.9cm] (fbox)
      {\textbf{0.835}\\{\scriptsize adjusted F1}\\[2pt]
       {\scriptsize 1{,}500 pages}};

    \draw[flow, cvprblue!70] (doc.east) to[out=35,in=180] (qbox.west);
    \draw[flow, orange!75]   (doc.east) to[out=-35,in=180] (lbox.west);
    \foreach \a/\b in {qbox/pbox, pbox/ebox, ebox/sbox}
      \draw[flow, cvprblue!70] (\a.east) -- (\b.west);
    \foreach \a/\b in {lbox/tbox, tbox/mbox, mbox/fbox}
      \draw[flow, orange!75] (\a.east) -- (\b.west);

    \begin{scope}[on background layer]
      \node[lane, fill=cvprblue!4, fit=(qbox)(sbox),
        inner sep=6pt] (qaband) {};
      \node[lane, fill=orange!4, fit=(lbox)(fbox),
        inner sep=6pt] (layband) {};
    \end{scope}
    \node[lanelbl, text=cvprblue!75, above=1pt of qaband.north west,
      anchor=south west] {QA track \;\textbar\; 1{,}356 questions};
    \node[lanelbl, text=orange!75!black, below=1pt of layband.south west,
      anchor=north west] {Layout track \;\textbar\; 1{,}500 annotated pages};
  \end{tikzpicture}}
  \caption{\textbf{Overview of \rdb}, a two-track benchmark over real regulated
  documents. \emph{Top (QA track):} each document yields hard field-level
  questions paired with a typed \texttt{gold\_dict}; every system's parsed
  markdown is read by a single extraction LLM and scored per field and per
  question with type-tolerant matching. \emph{Bottom (layout track):} the same
  document distribution is annotated with bounding boxes under a nine-class
  public taxonomy and scored with an adjacency-aware matcher. Both tracks run
  under one harness, and we report accuracy alongside per-page cost and
  cache-busted latency.}
  \label{fig:teaser}
\end{figure*}
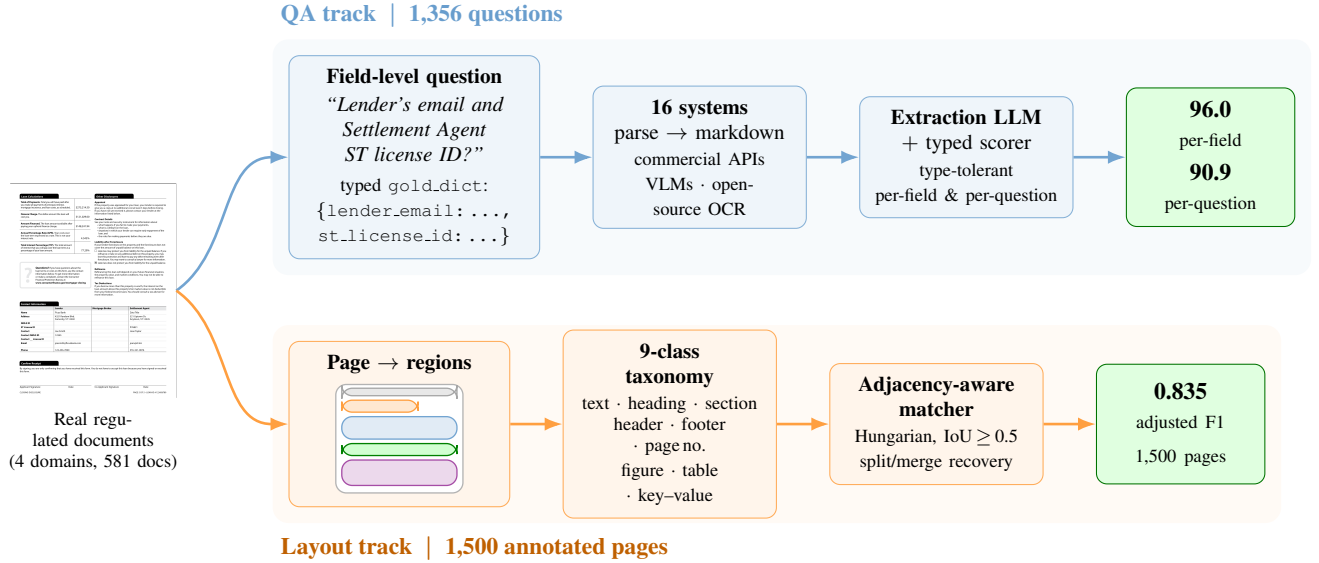

Automated document parsing--converting a PDF or scanned page into structured,
machine-readable content--underpins a growing class of enterprise workflows.
In many regulated industries such as insurance, financial lending, and healthcare, parsers increasingly feed autonomous agents that must extract the \emph{exact} value of a specific field (the cash-to-close figure on a Closing Disclosure, a premium on an ACORD form, a
quantity on a bill of lading, or a dosage on a clinical record) and then act
on it. In these domains, agents reading erroneously parsed documents can have massive financial and legal consequences. Additionally, parsers are increasingly used to convert documents into context for complex agent-based systems. An automated parsing solution therefore must represent all the information in a given document (including elements such as checkboxes, charts, handwritten text, and tables) accurately and preserve the underlying document structure.

Despite this, the benchmarks used to compare parsing systems are a poor proxy
for real-world performance. We observed three primary gaps: \textbf{(1) Unrealistic documents.}
Popular OCR and document-understanding benchmarks draw heavily on academic
papers, books, and born-digital web content with clean, predictable layouts.
Real regulated documents are messier: dense fillable forms, checkbox grids,
multi-column tables, handwriting, stamps, redactions, and scanner artifacts.
\textbf{(2) The wrong metric.} Many benchmarks score a parser by string- or
markdown-level similarity between its full-page output and a reference
transcription. A parser can score well on such metrics while still misreading
the single field a downstream system depends on, and a parser can be penalized
for benign formatting differences that do not affect any extracted value.
\textbf{(3) Single-number reporting.} A leaderboard with one aggregate score
hides where systems actually differ: which domains, which structural
features, and at what cost and latency.

We introduce \rdb \footnote{Dataset is available on hugging face: \url{https://huggingface.co/collections/Extend-AI/realdoc-benchmarks} and evaluation harness at \url{https://github.com/extend-hq/realdoc-bench}} to close these gaps. \rdb is built from real
regulated documents, and evaluates parsers on the task that matters in practice:
retrieving correct \emph{field values}. It comprises two complementary tracks
(\cref{fig:teaser}).

\paragraph{QA track.}
1356 field-level questions over 581 real documents in four domains
(mortgage, finance, supply chain, and medical/healthcare). Each question
targets a structurally hard extraction (a value bound to a row and column, a
checkbox state, a figure caption, a blank field that must be reported as
empty) and is paired with a typed \texttt{gold\_dict} mapping answer keys to
expected values. A parser's text output is run through a fixed extraction LLM and
then scored \emph{per field} (the fraction of keys whose value is correct) and
\emph{per question} (the fraction of questions for which \emph{every} key is
correct). This decouples the comparison from incidental formatting: only the
extracted values are judged, and they are judged identically for every system.

\paragraph{Layout track.}
1{,}500 human-verified page images annotated with COCO-style bounding boxes
under a nine-class public taxonomy (text, heading, section heading, header,
footer, page number, figure, table, key--value). Predictions are scored with a
Hungarian one-to-one matcher gated at IoU${\ge}0.5$ with adjacency-aware
split/merge recovery, reported as both a strict and an adjusted F1. Crucially,
ground truth and every model prediction are collapsed to the same public
taxonomy, so no system is judged on a private label set\footnote{\rdb is authored by the team at Extend. Our own systems are among the strongest
on the benchmark; we report their results factually alongside every other
system, hold all systems to an identical extraction-and-scoring protocol, and
release the datasets, adapters, and harness so that any result can be
reproduced or contested.}.

\paragraph{Contributions.}
\begin{itemize}
  \item \textbf{A real, field-level QA benchmark.} 581 regulated
    documents and 1{,}356 questions with typed gold answers and structural
    capability tags, scored with value-level metrics that ignore benign
    formatting differences.
  \item \textbf{A paired layout benchmark.} 1500 verified pages on a nine-class
    taxonomy with an adjacency-aware matcher and a symmetric,
    auditable fairness protocol across vendors.
  \item \textbf{An open-sourced uniform, reproducible evaluation of eighteen systems} (commercial
    parsing APIs, general-purpose VLMs, and open-source OCR models), reporting
    accuracy together with per-page cost and cache-busted latency.
  \item \textbf{Finegrained analysis of parser performance}: per-domain and
    per-capability breakdowns and a cost/latency trade-off analysis that exposes
    distinct operating points rather than one dominant system.
\end{itemize}

\section{Related Work}
\label{sec:related}

\paragraph{Document VQA and key-information extraction.}
A long line of work evaluates reading and reasoning over documents.
DocVQA~\cite{mathew2021docvqa} and InfographicVQA~\cite{mathew2022infographicvqa}
pose natural-language questions over single document images;
FUNSD~\cite{jaume2019funsd}, CORD~\cite{park2019cord}, and the SROIE
task~\cite{huang2019sroie} target form understanding and receipt
key-information extraction. These datasets established the value of
field-level supervision, but are dominated by a single document genre (scanned
forms, receipts) and relatively short answers. \rdb keeps the field-level
supervision but draws from full multi-page regulated documents across four
industries, and pairs each question with a \emph{typed} gold dictionary so that
numeric, date, boolean, and enumerated answers are scored by value rather than
by string overlap.

\paragraph{OCR and full-page parsing benchmarks.}
Recent benchmarks evaluate end-to-end page parsing into markdown or structured
text. OmniDocBench~\cite{ouyang2024omnidocbench} provides diverse page types
with fine-grained annotations and edit-distance--style metrics;
OCRBench~\cite{liu2024ocrbench} aggregates OCR-centric tasks for multimodal
models. Such page-level similarity metrics reward faithful transcription but
do not isolate whether a downstream-critical \emph{value} was read correctly,
and they can penalize benign formatting choices. \rdb{}'s QA track
instead scores the values a downstream agent would actually query, using
type-tolerant equality and conservative fuzzy matching, and reports a strict
per-question metric that requires \emph{all} fields of a question to be correct.

\paragraph{Granular and rule-based parsing benchmarks.}
A recent line of benchmarks shares our diagnosis, that page-level similarity is
the wrong signal, and replaces it with large numbers of small,
machine-checkable assertions. \textbf{olmOCR-Bench}~\cite{olmocrbench} comprises
$7{,}010$ pass/fail unit tests over $1{,}402$ PDFs drawn largely from arXiv, the
Internet Archive, and the Library of Congress; each test checks an unambiguous
fact about a page (a short text segment is present, a header/footer is
\emph{absent}, reading order is preserved, a table cell or formula is correct).
It rejects holistic, LLM-as-judge similarity scoring, but still relies on fuzzy
string matching for its text-presence tests. Two design choices narrow its
relevance to regulated extraction: its ``text absence'' tests encode the
editorial judgment that headers, footers, and page numbers are noise to be
\emph{dropped} (itself a parsing opinion, not a universal ground truth), and
nearly half of its tests are mathematical-formula checks, reflecting its
arXiv-heavy provenance.
\textbf{ParseBench}~\cite{parsebench} targets the enterprise setting closest to
ours: roughly $2{,}000$ human-verified pages with over $167{,}000$ rules across
five capability dimensions (tables, charts, content faithfulness, semantic
formatting, and visual grounding), and reports a fragmented landscape in which
no single system dominates every dimension (LlamaParse Agentic is its strongest
overall at $84.9\%$).

\rdb is complementary to both (\cref{tab:bench-compare}). Like olmOCR-Bench and
ParseBench, we reject surface-similarity scoring; unlike them, our unit of supervision is the \emph{typed value of a named field}, the object a downstream agent actually queries rather than a free-text fact (olmOCR-Bench) or a formatting/structure rule (ParseBench). Typed named fields allow us to measure accuracy across different value types (for example, strings, integers, and currency). We also report a stricter per-question metric that requires \emph{every} field of a record to be correct simultaneously.  Our documents are real regulated forms across four
industries (with privacy-preserving synthesis applied only to sensitive
medical and tax content; \cref{sec:construction:sourcing}). While
olmOCR-Bench skews academic/web, ParseBench overlaps our
enterprise focus. However, the metrics focus more on parse \emph{quality} dimensions (which could be subjective choices) rather than
extracted values. Finally, \rdb uniquely pairs its QA track with a
bounding-box \emph{layout} track on a public taxonomy, and reports per-page cost
and cache-busted latency alongside accuracy. Encouragingly, the two benchmarks
agree on the ordering of overlapping systems: LlamaParse Agentic is the
strongest non-Extend system on \rdb as well.

\begin{table*}[t]
  \centering
  \small
  \resizebox{\textwidth}{!}{
  \begin{tabular}{lllll}
    \toprule
    Benchmark & Scale & Documents & Unit of supervision & Primary metric \\
    \midrule
    DocVQA~\cite{mathew2021docvqa} & 50k Q / 12k imgs & mixed scanned & NL question $\rightarrow$ short answer & ANLS \\
    OmniDocBench~\cite{ouyang2024omnidocbench} & $\sim$1k pages & diverse (academic-heavy) & full-page parse & edit distance / TEDS \\
    olmOCR-Bench~\cite{olmocrbench} & 7{,}010 tests / 1{,}402 PDFs & arXiv, archive, LoC & page-level pass/fail facts & unit-test pass rate \\
    ParseBench~\cite{parsebench} & $\sim$2k pages / 167k rules & enterprise & structure/quality rules (5 dims) & rule pass rate \\
    \textbf{\rdb (ours)} & 1{,}356 Q / 581 docs $+$ 1.5k layout pages & real regulated (4 domains) & typed field value $+$ layout box & per-field/-question acc., adj.\ F1 \\
    \bottomrule
  \end{tabular}}
  \caption{\rdb in context. Prior benchmarks either score full-page similarity
  (OmniDocBench) or large sets of free-text/structure assertions (olmOCR-Bench,
  ParseBench). \rdb supervises the \emph{typed value of a named field}, adds a
  paired layout track on a public taxonomy, and reports cost and latency.}
  \label{tab:bench-compare}
\end{table*}

\paragraph{Layout analysis datasets.}
PubLayNet~\cite{zhong2019publaynet}, DocLayNet~\cite{pfitzmann2022doclaynet},
and DocBank~\cite{li2020docbank} are standard for document layout detection,
but skew toward scientific and born-digital documents and use
taxonomies tuned to those genres. \rdb{}'s layout track targets the same
regulated-document distribution as its QA track, uses a nine-class
taxonomy chosen for downstream parsing (including a \texttt{key--value} class),
and scores predictions with an adjacency-aware matcher that credits correct
content split across box boundaries.

\paragraph{Parsing systems.}
The systems we evaluate include commercial document-AI services (Amazon
Textract~\cite{aws_textract}, Azure Document Intelligence~\cite{azure_di},
Reducto~\cite{reducto}, and LlamaParse~\cite{llamaparse}), general-purpose
vision-language models used as parsers, and a growing set of open-source OCR
models including olmOCR~\cite{poznanski2024olmocr},
Docling~\cite{auer2024docling}, Nanonets-OCR~\cite{nanonetsocr},
dots.ocr~\cite{dotsocr}, PaddleOCR-VL~\cite{paddleocrvl}, and
Chandra~\cite{chandra}. Prior
comparisons of these systems are typically vendor-published and use
inconsistent metrics. \rdb{} places all of them under one harness with a single
extraction LLM, a single scorer, and disclosed cost and latency protocols.

\paragraph{Why a new benchmark?}
Four gaps motivated \rdb, none of which is fully
addressed by any single existing benchmark. \textbf{(1) The unit of evaluation
is still not the field value.} olmOCR-Bench checks whether spans of \emph{text}
appear, are absent, or are ordered correctly; ParseBench checks parse-quality
\emph{rules}; OmniDocBench scores page similarity. A system can satisfy all of
these while still binding the wrong number to a field in a multi-column grid or
a checkbox block, which is precisely the failure that breaks a downstream
extraction pipeline. We wanted to score the \emph{typed value of a named field},
and to require, via a strict per-question metric, that an entire record be
correct at once. \textbf{(2) The document distribution is not regulated nor
real.} Academic and web PDFs do not exhibit the handwriting, checkbox grids,
stamps, redactions, and dense fillable forms that dominate mortgage, insurance,
logistics, and clinical workflows and that drive real error modes; we needed a
corpus built from such documents, spanning four industries.
\textbf{(3) QA and layout are not evaluated on the same distribution.}
No prior benchmark pairs field-level QA with a bounding-box layout track
over the \emph{same} regulated documents, so one cannot ask whether a
QA error traces back to a layout-detection failure; \rdb{} provides both
tracks under one taxonomy and one document distribution. \textbf{(4) Operational
cost is unreported.} Deployment decisions hinge on dollars and latency per page,
not accuracy alone, yet these axes are largely absent from existing
leaderboards; \rdb reports cost and latency alongside accuracy.
Building the benchmark ourselves also let us control and \emph{publish} the
provenance, gold-validation, taxonomy mappings, and per-vendor adapters, so that
every number can be audited or reproduced.

\section{Benchmark Construction}
\label{sec:construction}

\rdb has two tracks that share a design philosophy: real regulated documents,
public taxonomies, and supervision at the granularity a downstream agent
queries, but differ in their unit of evaluation. The QA track
(\cref{sec:construction:sourcing,sec:construction:qa}) supervises field
\emph{values}; the layout track (\cref{sec:construction:layout}) supervises
region \emph{geometry and type}.

\subsection{Document sourcing and domains}
\label{sec:construction:sourcing}

The QA track draws from a curated pool of real documents organized by
document type within four domains: \textbf{mortgage} (Closing
Disclosures, FNMA~710, TREC contracts, URLA), \textbf{finance} (ACORD
insurance forms, SEC filings, tax forms), \textbf{supply chain} (bills of
lading, commercial invoices), and \textbf{medical/healthcare} (patient
intake forms, clinical consent forms, case-report forms). Documents are sourced
to reflect the genres encountered in production parsing workloads rather than
the clean layouts common in academic corpora.

Sourcing applies an explicit quality gate. We pass candidate documents through
rejection and quarantine stages that exclude low-fidelity or inappropriate
material, for example pages bearing confidentiality stamps or documents whose
content does not match the claimed document type. After
selection, documents are normalized to single-page PDFs and rendered to images
for the question-generation and verification stages.

\paragraph{Privacy-preserving sourcing for sensitive domains.}

Most of the data has been sourced from public governmental archives to prevent copyright
infringement, and avoid leaking private data. Specifically, medical/healthcare and tax documents can carry protected personal information.
Where redistributable real instances exist, for example tax filings from
historical public archives, we use them directly. Otherwise we synthesize
documents by populating \emph{real, blank form templates} with
fictitious-persona data, preserving the authentic layout while fabricating only
the field \emph{contents}. A vision-language fidelity check confirms the
template is filled rather than altered, and the persona values serve as gold by
construction. We detail this pipeline in \cref{app:synthesis}. 

The final QA bank contains 581 documents and 1356 questions.
\cref{tab:domains} reports the distribution across domains, which is
intentionally uneven: it tracks the relative prevalence and difficulty of each
genre rather than enforcing an artificial balance.

\begin{table}[t]
  \centering
  \small
  \setlength{\tabcolsep}{4pt}
  \begin{tabular}{lrrrr}
    \toprule
    Domain & Questions & Docs & Set A & Set B \\
    \midrule
    Mortgage           & 478 (35.2\%) & 222 (38.2\%) & 224 & 254 \\
    Finance            & 378 (27.9\%) & 172 (29.6\%) & 378 & 0 \\
    Supply chain       & 319 (23.5\%) & 113 (19.4\%) & 76  & 243 \\
    Medical/healthcare & 181 (13.3\%) & 74 (12.7\%)  & 169 & 12 \\
    \midrule
    \textbf{Total}     & \textbf{1{,}356} & \textbf{581} & 847 & 509 \\
    \bottomrule
  \end{tabular}
  \caption{QA-track composition by domain. Set~A and Set~B count the
  number of questions contributed from each of two independently curated
  document collections.}
  \label{tab:domains}
\end{table}

\subsection{QA-track question construction}
\label{sec:construction:qa}

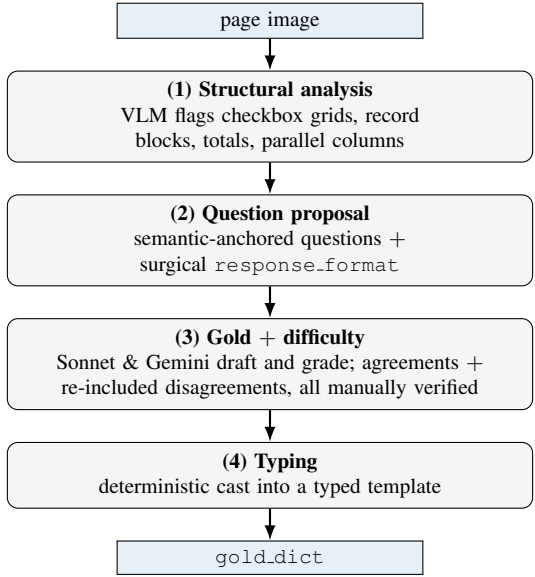
\begin{figure}[t]
  \centering
  \begin{tikzpicture}[
    font=\footnotesize,
    stage/.style={draw, rounded corners, align=center, inner sep=4pt,
      text width=0.80\linewidth, fill=gray!8},
    io/.style={draw, align=center, inner sep=3pt, text width=0.46\linewidth,
      fill=cvprblue!12},
    arr/.style={-{Latex[length=2mm]}, thick},
    node distance=4.2mm,
  ]
    \node[io] (img) {page image};
    \node[stage, below=of img] (s1)
      {\textbf{(1) Structural analysis}\\ VLM flags checkbox grids, record
      blocks, totals, parallel columns};
    \node[stage, below=of s1] (s2)
      {\textbf{(2) Question proposal}\\ semantic-anchored questions $+$ surgical
      \texttt{response\_format}};
    \node[stage, below=of s2] (s3)
      {\textbf{(3) Gold $+$ difficulty}\\ Sonnet \& Gemini draft and grade;
      agreements $+$ re-included disagreements, all manually verified};
    \node[stage, below=of s3] (s4)
      {\textbf{(4) Typing}\\ deterministic cast into a typed template};
    \node[io, below=of s4] (gd) {\texttt{gold\_dict}};
    \draw[arr] (img) -- (s1);
    \draw[arr] (s1) -- (s2);
    \draw[arr] (s2) -- (s3);
    \draw[arr] (s3) -- (s4);
    \draw[arr] (s4) -- (gd);
  \end{tikzpicture}
  \caption{\textbf{Question-generation pipeline.} (1)~A VLM analyzes the page
  image for structural challenges; (2)~it proposes hard, semantically anchored
  questions with a surgical \texttt{response\_format}; (3)~two models (Sonnet and
  Gemini) independently draft gold and grade difficulty; agreements plus a
  difficulty-weighted portion of disagreements are kept and every retained item
  is manually verified; (4)~a deterministic step types the answer into a
  \texttt{gold\_dict}.}
  \label{fig:qa-pipeline}
\end{figure}

Each question targets a structurally hard, content-level extraction and is
paired with a typed gold answer. Questions are produced by a multi-stage
pipeline (\cref{fig:qa-pipeline}) and then converted into a machine-checkable
form.

\paragraph{(1) Structural analysis.}
A vision-language model (Gemini~3 Flash) inspects each page image and
identifies the structural features that make extraction difficult: checkbox
grids, repeated record blocks, field--value pairs, totals, and parallel
columns. This analysis conditions question generation on the actual layout
challenges present on the page.

\paragraph{(2) Question proposal.}
For each page the model proposes hard, structural questions under a strict
specification. Questions must (i) be anchored by \emph{semantic} labels (e.g.\
``in the signatory block for `Signature on behalf of packer'\,'') rather than just
visual position; (ii) focus on structure (row/column binding, parallel
columns, checkbox state) rather than headers, footers, or page furniture; and
(iii) declare a surgical \texttt{response\_format} that specifies every answer
key and delimiter, e.g.\ \texttt{automobile\_premium=<number>}. Gold answers are
proposed as flat strings of the form \texttt{key1=value1; key2=value2}.

\paragraph{(3) Gold validation and difficulty grading.}
Gold answers are drafted independently by two different models, Claude Sonnet
and Gemini, working from the page image, and each question is graded for
difficulty (easy / intermediate / hard). Questions on which the two models agree
are retained directly. Rather than discarding every disagreement (which is
concentrated in the intermediate and hard tiers and is precisely the items that
separate parsers), we add a portion of them back into the bank.  Every retained
question, whether from an agreement or a re-included disagreement, is then
manually verified against the source page by a human annotator who sets the gold
value. This cross-model drafting, difficulty-aware re-inclusion, and full manual
verification, rather than a single automated re-derivation pass, is what makes
the gold reliable while keeping genuinely hard items in the bank.

\paragraph{(4) Typing (jsonification).}
A deterministic step converts each \texttt{response\_format} into a typed
template and each free-form gold string into a typed \texttt{gold\_dict}.
Placeholders map to a small type vocabulary (integer, number, date
in \texttt{YYYY-MM-DD} form, boolean, and enumerations \texttt{<one of: a | b>}),
and the gold value is parsed into the corresponding JSON type (so \texttt{"12800"}
becomes the integer \texttt{12800}, \texttt{"yes"} becomes \texttt{true}).
Typing is what allows scoring to be type-aware and tolerant of formatting
(\cref{sec:method:qa}).

\paragraph{Question record.}
A bank entry pairs a natural-language question with its typed answer and
metadata, illustrated below (abbreviated):
\begin{quote}\small\ttfamily
question\_id, source\_file, domain,\\
question, response\_format, gold\_answer,\\
gold\_dict, capabilities, origin, set
\end{quote}
The \texttt{capabilities} field tags each question with the structural skills it
exercises (e.g.\ \texttt{field\_value\_pairing}, \texttt{multi\_column\_grid},
\texttt{checkbox\_state}, \texttt{chart\_value}). These tags are grouped into a
small number of capability buckets (checkboxes, tables, forms/key--value,
multi-column, charts and figures, handwriting, blank-field detection, and
small text), enabling per-capability error analysis in addition to per-domain
breakdowns.

\begin{figure}[t]
  \centering
  \setlength{\fboxsep}{6pt}
  \noindent\fbox{\parbox{0.95\linewidth}{\small
    \textbf{Domain:} finance \quad\textbf{Doc:} ACORD 125/126 (commercial
    insurance application)\\[2pt]
    \textbf{Question:} \textit{In the PRIOR CARRIER INFORMATION (continued)
    table, for the year 201 entry with an expiration date of 12/31/2024, what
    is the premium for the automobile category?}\\[2pt]
    {\ttfamily response\_format:automobile\_premium=<number>}\\
    {\ttfamily gold\_answer:\ \ \ \ automobile\_premium=12800}\\
    {\ttfamily gold\_dict:\ \ \ \ \ \ \ \{"automobile\_premium": 12800\}}\\[2pt]
    \textbf{capabilities:} {\small field\_value\_pairing, multi\_column\_grid,
    row\_binding, table\_structure}
  }}
  \caption{\textbf{Example QA item.} A real item from the finance (ACORD)
  domain. The answer is bound to a specific row and column of a multi-column
  premium table, the kind of structural extraction that page-level similarity
  metrics fail to isolate. The typed \texttt{gold\_dict} lets scoring compare
  the integer value rather than the surface string. Additional per-domain
  examples appear in \cref{app:examples}.}
  \label{fig:example-qa}
\end{figure}

\subsection{Layout-track annotation and taxonomy}
\label{sec:construction:layout}

The layout track contains 1{,}500 page images with COCO-style bounding-box
annotations. Each annotation carries a box, a category, and standard COCO
fields; images are a mix of rendered and scanned pages.

\begin{figure}[t]
  \centering
  \begin{minipage}[c]{0.45\linewidth}
    \centering
    \begin{tikzpicture}[x=3.9cm, y=-5.05cm, line width=0.4pt]
      \node[anchor=north west, inner sep=0] at (0,0)
        {\includegraphics[width=3.9cm]{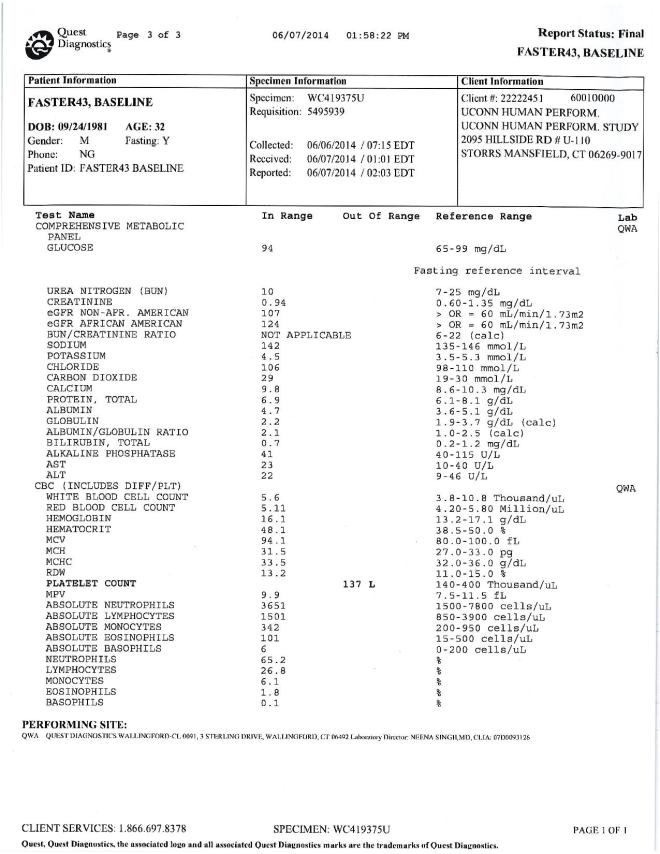}};
      \draw[gray!55] (0,0) rectangle (1,1);
      \foreach \cls/\x/\y/\w/\h in {%
        orange/0.036/0.030/0.132/0.040,
        olive/0.176/0.035/0.097/0.012,
        teal/0.412/0.035/0.086/0.012,
        teal/0.523/0.036/0.098/0.009,
        violet/0.814/0.032/0.162/0.014,
        blue/0.783/0.054/0.192/0.014,
        violet/0.035/0.085/0.941/0.158,
        red/0.052/0.247/0.912/0.580,
        cyan/0.032/0.842/0.161/0.011,
        gray/0.030/0.857/0.773/0.009,
        brown/0.032/0.963/0.250/0.013,
        brown/0.414/0.965/0.177/0.012,
        olive/0.868/0.966/0.080/0.011,
        brown/0.030/0.982/0.724/0.012}{%
        \fill[\cls, opacity=0.16] (\x,\y) rectangle ({\x+\w},{\y+\h});
        \draw[\cls, opacity=0.9] (\x,\y) rectangle ({\x+\w},{\y+\h});
      }
    \end{tikzpicture}
  \end{minipage}\hfill
  \begin{minipage}[c]{0.5\linewidth}
    \footnotesize\setlength{\tabcolsep}{4pt}
    \begin{tabular}{@{}ll@{\hspace{6pt}}ll@{}}
      {\color{gray}\rule{7pt}{7pt}}   & text    & {\color{olive}\rule{7pt}{7pt}}  & page number \\
      {\color{blue}\rule{7pt}{7pt}}   & heading & {\color{orange}\rule{7pt}{7pt}} & figure \\
      {\color{cyan}\rule{7pt}{7pt}}   & section & {\color{red}\rule{7pt}{7pt}}    & table \\
      {\color{teal}\rule{7pt}{7pt}}   & header  & {\color{violet}\rule{7pt}{7pt}} & key--value \\
      {\color{brown}\rule{7pt}{7pt}}  & footer  &                                 &            \\
    \end{tabular}
  \end{minipage}
  \caption{\textbf{Example layout annotation.} A real human-verified page
  (medical domain) with the nine-class bounding boxes overlaid on the source
  document. Dense forms produce large adjacent \texttt{key\_value} and
  \texttt{table} regions, motivating the adjacency-aware matcher of
  \cref{sec:method:layout}.}
  \label{fig:example-layout}
\end{figure}
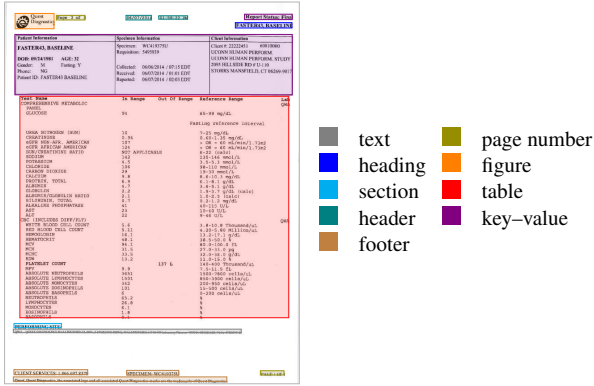

\paragraph{Public nine-class taxonomy.}
Production layout taxonomies are large and partly vendor-specific. To make
cross-system comparison fair and auditable, \rdb exposes a single nine-class
public taxonomy: \texttt{text}, \texttt{heading}, \texttt{section\_heading},
\texttt{header}, \texttt{footer}, \texttt{page\_number}, \texttt{figure},
\texttt{table}, and \texttt{key\_value}.

\paragraph{Annotation pipeline.}
Annotations are produced by a model-assisted human verification flow: a layout
model emits candidate regions which annotators review and correct in a
labeling tool, rather than annotating each page from scratch. This keeps the
ground truth aligned with the public taxonomy while retaining a human in the
loop for every page. Pages are rendered at high resolution for annotation and
selected to be content-rich (tables, forms, key--value, and figures) rather
than blank or cover pages, so the track stresses the structures that matter for
downstream parsing.

\section{Evaluation Methodology}
\label{sec:method}

Every system is evaluated under one harness. For the QA track, a
parser's text output is passed through a fixed extraction LLM and scored against the
typed gold dictionary (\cref{sec:method:qa}). For the layout track, predictions
are normalized to the nine-class taxonomy and matched against ground truth with
an adjacency-aware scorer (\cref{sec:method:layout}). We additionally report a
disclosed cost and latency protocol (\cref{sec:method:cost}).

\subsection{QA-track scoring}
\label{sec:method:qa}

\paragraph{Two-stage protocol.}
A parser produces a text (markdown) rendering of a document; it does not see the
questions. For each (question, parser) pair, a fixed extraction LLM
(Gemini~3 Flash) is given the parser's markdown, the question, and the typed
answer template, and is instructed to return a JSON object with exactly the
template's keys and types, copying values verbatim from the markdown and using
no outside knowledge. Redaction markers map to null; stray OCR
character-spacing (e.g.\ ``\texttt{0 . 0 0}'') is collapsed. Decoupling
extraction from parsing is what lets a single protocol compare systems with
very different output styles: every parser is read by the \emph{same} extraction LLM,
so differences in markdown formatting do not advantage or penalize any system.
The same extraction LLM and prompt are used for all parsers.

\paragraph{Type-aware comparison.}
The extracted JSON is compared to \texttt{gold\_dict} key by key. Comparison is
deliberately tolerant of differences that do not change a value:
\begin{itemize}
  \item \textbf{Deep equality} normalizes numbers against numeric strings with
    currency, percent, and thousands separators (\texttt{"\$1,234"}${=}$\,$1234$),
    tolerates smart-quote and dash variants, strips markdown styling
    (\texttt{**}, \texttt{\_\_}, backticks), collapses OCR character spacing,
    and unwraps a single-element list around a dict.
  \item \textbf{Conservative fuzzy match} applies only to longer multi-word
    string fields, accepting a match above a high similarity threshold
    ($\ge 92$); short fields require exact (normalized) equality.
\end{itemize}
Each field yields a boolean; a question is counted correct only if
\emph{every} one of its keys matches.

\paragraph{Metrics.}
We report two accuracies. \textbf{Per-field accuracy} is the fraction of all
answer keys (across all questions) whose value is correct. \textbf{Per-question
accuracy} is the fraction of questions for which all keys are correct: a
strict metric that reflects whether a downstream agent would get a fully
correct record. Both are reported overall, per domain, and per capability
bucket. We report 95\% confidence intervals from a \emph{document-clustered}
bootstrap (\cref{app:ci}).

\subsection{Layout-track scoring}
\label{sec:method:layout}

\paragraph{Normalization.}
Each system's native layout output is mapped into the nine-class public
taxonomy by a per-vendor normalizer applied once at load time. The mappings are
published so that every vendor's collapse can be audited.

\paragraph{Adjacency-aware matching.}
Predictions are matched to ground-truth boxes by a Hungarian one-to-one
assignment that minimizes a cost combining localization and type error,
\begin{equation}
  c(p, g) = \bigl(1 - \mathrm{IoU}(p, g)\bigr) + 0.25 \cdot \mathbf{1}[\,\tau(p) \neq \tau(g)\,],
\end{equation}
where $\tau(\cdot)$ is the predicted/true class, gated so that only pairs with
$\mathrm{IoU} \ge 0.5$ may match. Because real documents are often segmented
differently by different systems (one model emits a single table region where
another emits several adjacent ones), a second pass attempts \emph{adjacency
recovery}: unmatched same-type boxes that are spatially adjacent (small gap),
aligned along the perpendicular axis, and whose merge clears the IoU gate are
combined and rematched. This yields two scores per page: a \textbf{strict} F1
from the one-to-one pass alone, and an \textbf{adjusted} F1 after adjacency
recovery. We report micro, macro, and per-class F1, with precision and recall.
Note that not all models we analyze output all layout types.

\paragraph{Symmetric fairness.}
Ground truth and predictions are scored on the same nine-class taxonomy. A system whose output vocabulary structurally lacks a class (for instance a detector with no \texttt{key\_value} head) simply scores zero on
that class: it incurs the false negatives present in the ground truth but is
not additionally penalized, because both sides of the gap come from the same
matcher. The taxonomy collapse happens once, at dataset build time, and is
applied identically to both sides. Penalizing a system for a class its output
vocabulary structurally lacks is a known difficulty for layout benchmarks; we do
not eliminate it, but make it transparent and auditable by scoring every system
on the same published taxonomy and disclosing each vendor's collapse.

\subsection{Cost and latency protocol}
\label{sec:method:cost}

\paragraph{Cost.}
We report per-page list price with no enterprise discounts. Credit-priced
systems are converted to dollars per page using each vendor's published
credit rate and observed credit usage; flat-rate services use their per-page
price. Where a system has a base and an ``agentic'' mode, the surcharge is
taken from the vendor's own per-call credit breakdown.

\paragraph{Latency.}
Latency is the median of five cache-busted solo calls per system on the same
one-page document, measured with all systems running concurrently (each
system's own calls remain serial). To defeat provider-side parse caches, each
call appends a unique nonce after the PDF's \texttt{\%\%EOF} marker, giving every
request a fresh content hash. Reported latencies therefore reflect cold parses;
sustained throughput and multi-page documents may scale differently per
provider.

\section{Results}
\label{sec:results}

We evaluate eighteen systems on the QA track: commercial parsing APIs
(Extend Performance~v2, Extend~v1, LlamaParse and LlamaParse Agentic, Reducto
and Reducto Agentic, Azure Document Intelligence, Amazon Textract), two
general-purpose models used as parsers (Gemini~3.5 Flash~\cite{gemini} and
Claude Opus~4.8~\cite{claude}), and eight open-source OCR systems
(Chandra-2~\cite{chandra}, olmOCR-2~\cite{olmocrbench},
Nanonets-OCR-s~\cite{nanonetsocr}, Docling~\cite{auer2024docling},
dots.ocr~\cite{dotsocr} in two configurations, Nemotron Nano
V2 VL~\cite{nemotronnanovl}, PaddleOCR-VL~\cite{paddleocrvl}, and Paddle
v3~\cite{paddleocr3}). The layout track is evaluated on five
systems for which a layout-region output is available.

\subsection{QA leaderboard}
\label{sec:results:leaderboard}

\cref{tab:leaderboard} reports per-field and per-question accuracy on the full
1{,}356-question bank. Per-field accuracy is over $3{,}742$ answer keys.

\begin{table}[t]
  \centering
  \small
  \setlength{\tabcolsep}{4pt}
  \begin{tabular}{lrr}
    \toprule
    System & Per-field & Per-question \\
    \midrule
    \textbf{Extend Performance v2} & \textbf{96.0}\,{\scriptsize$\pm$0.8} & \textbf{90.9}\,{\scriptsize$\pm$1.6} \\
    Claude Opus 4.8 (non-thinking)~\cite{claude} & 93.0\,{\scriptsize$\pm$1.4}  & 87.1\,{\scriptsize$\pm$2.1} \\
    LlamaParse (Agentic)           & 92.2\,{\scriptsize$\pm$1.5} & 84.5\,{\scriptsize$\pm$2.2} \\
    Reducto (Agentic)              & 91.4\,{\scriptsize$\pm$1.7} & 83.8\,{\scriptsize$\pm$2.3} \\
    Extend v1                      & 90.8\,{\scriptsize$\pm$1.7} & 82.5\,{\scriptsize$\pm$2.4} \\
    Gemini 3.5 Flash               & 89.3\,{\scriptsize$\pm$1.8} & 82.2\,{\scriptsize$\pm$2.3} \\
    LlamaParse                     & 89.2\,{\scriptsize$\pm$2.0} & 80.8\,{\scriptsize$\pm$2.5} \\
    Azure DI                       & 89.1\,{\scriptsize$\pm$1.7} & 79.6\,{\scriptsize$\pm$2.5} \\
    Reducto                        & 88.7\,{\scriptsize$\pm$2.0} & 80.5\,{\scriptsize$\pm$2.5} \\
    AWS Textract                   & 70.7\,{\scriptsize$\pm$2.7} & 54.0\,{\scriptsize$\pm$2.9} \\
    \midrule
    \multicolumn{3}{l}{\textit{Open-source}}\\
    Chandra-2                      & 86.2\,{\scriptsize$\pm$2.4} & 78.1\,{\scriptsize$\pm$2.7} \\
    olmOCR-2                       & 79.5\,{\scriptsize$\pm$2.6} & 67.9\,{\scriptsize$\pm$3.0} \\
    Nanonets-OCR-s                 & 77.4\,{\scriptsize$\pm$3.6} & 68.2\,{\scriptsize$\pm$3.2} \\
    Docling                        & 71.2\,{\scriptsize$\pm$2.9} & 54.7\,{\scriptsize$\pm$3.2} \\
    dots.ocr (2-stage)             & 70.6\,{\scriptsize$\pm$3.6} & 61.4\,{\scriptsize$\pm$3.5} \\
    dots.ocr (single-shot)         & 65.4\,{\scriptsize$\pm$4.0} & 57.2\,{\scriptsize$\pm$3.6} \\
    Paddle-OCR-VL~\cite{paddleocrvl} & 59.6\,{\scriptsize$\pm$4.0} & 48.5\,{\scriptsize$\pm$3.6} \\
    Paddle v3~\cite{paddleocr3}    & 59.2\,{\scriptsize$\pm$3.2} & 43.1\,{\scriptsize$\pm$3.2} \\
    Nemotron Nano V2 VL            & 41.6\,{\scriptsize$\pm$3.4} & 29.3\,{\scriptsize$\pm$3.0} \\
    \bottomrule
  \end{tabular}
  \caption{Full QA leaderboard (1{,}356 questions, 581 documents). Accuracies in
  percent, with 95\% document-clustered bootstrap confidence intervals shown as
  half-widths ($B{=}10{,}000$; \cref{sec:method:qa}); the full intervals are
  plotted in \cref{fig:forest}. AWS Textract is a commercial service but lands among the
  open-source tier on this benchmark.}
  \label{tab:leaderboard}
\end{table}

Extend Performance~v2 leads on both metrics (96.0\% per-field, 90.9\%
per-question), $3.0$ points ahead of the next system per-field. The agentic
modes of LlamaParse and Reducto form a second tier, followed by a cluster of
single-pass commercial systems and Gemini~3.5 Flash within roughly one point of
one another (88.7--89.3\% per-field). The open-source field is led by Chandra-2
at 86.2\%/78.1\%, within striking distance of the weaker commercial systems,
while the gap between the strongest and weakest evaluated systems exceeds 50
points per-field. The strict per-question metric spreads systems further than
per-field: requiring \emph{all} keys to be correct drops most systems by
5--17 points relative to their per-field score, underscoring that
field-averaged accuracy can overstate end-to-end reliability.
The 95\% document-clustered bootstrap intervals (\cref{tab:leaderboard},
visualized in \cref{fig:forest}) make the tier structure explicit: Extend
Performance~v2's per-field interval is disjoint from every other system's,
whereas the single-pass commercial systems (Gemini, LlamaParse, Azure~DI, and
Reducto) form a cluster whose intervals overlap and are therefore not
statistically separable on this bank.

\subsection{Per-domain analysis}
\label{sec:results:domain}

\cref{tab:perfield-domain,tab:perq-domain} break the nine headline systems
down by domain.

\begin{table}[t]
  \centering
  \small
  \setlength{\tabcolsep}{4pt}
  \begin{tabular}{lrrrr}
    \toprule
    System & Fin. & Mort. & Supply & Med. \\
     & (378) & (478) & (319) & (181) \\
    \midrule
    \textbf{Extend\ v2} & \textbf{92.7} & \textbf{97.5} & \textbf{97.6} & \textbf{89.8} \\
    LlamaParse (Agentic)      & 86.1 & 95.1 & 94.7 & 82.6 \\
    Reducto (Agentic)         & 86.1 & 92.5 & 94.3 & 86.6 \\
    Extend v1                 & 86.9 & 94.9 & 89.9 & 82.9 \\
    Gemini 3.5 Flash          & 83.8 & 92.5 & 92.1 & 75.8 \\
    LlamaParse                & 83.2 & 92.9 & 90.3 & 80.7 \\
    Azure DI                  & 82.9 & 93.7 & 90.3 & 76.7 \\
    Reducto                   & 83.8 & 93.2 & 88.6 & 79.2 \\
    AWS Textract              & 68.4 & 71.6 & 77.0 & 46.9 \\
    \bottomrule
  \end{tabular}
  \caption{Per-field accuracy (\%) by domain (full bank). Question counts in
  parentheses.}
  \label{tab:perfield-domain}
\end{table}

\begin{table}[t]
  \centering
  \small
  \setlength{\tabcolsep}{4pt}
  \begin{tabular}{lrrrr}
    \toprule
    System & Fin. & Mort. & Supply & Med. \\
     & (378) & (478) & (319) & (181) \\
    \midrule
    \textbf{Extend\ v2} & \textbf{89.2} & \textbf{93.5} & \textbf{92.8} & \textbf{84.5} \\
    LlamaParse (Agentic)      & 79.4 & 87.7 & 88.7 & 79.6 \\
    Reducto (Agentic)         & 81.0 & 86.0 & 85.9 & 80.7 \\
    Extend v1                 & 80.4 & 88.3 & 78.7 & 78.5 \\
    Gemini 3.5 Flash          & 81.0 & 85.6 & 85.6 & 70.2 \\
    LlamaParse                & 76.7 & 87.0 & 77.4 & 79.0 \\
    Azure DI                  & 76.7 & 84.7 & 79.0 & 72.9 \\
    Reducto                   & 76.7 & 86.8 & 77.7 & 76.8 \\
    AWS Textract              & 54.8 & 54.6 & 58.3 & 43.1 \\
    \bottomrule
  \end{tabular}
  \caption{Per-question accuracy (\%) by domain (full bank).}
  \label{tab:perq-domain}
\end{table}

Several patterns emerge. \textbf{Mortgage is the easiest domain}: the top
systems sit within roughly five points of one another at 92--98\% per-field,
reflecting the clean fillable forms (Closing Disclosures, FNMA~710, TREC
contracts, URLA) that dominate it. \textbf{Medical is the hardest}: the best
systems reach only 86--90\% per-field and the weakest collapses to 47\%, driven
by handwritten case-report forms, autopsy reports, and consent forms with
checkbox grids. \textbf{Finance is the most discriminative} domain, with the
widest spread among strong systems (Extend~v2 at 92.7\% vs.\ Azure~DI at 82.9\%,
a $\sim$10-point range). \textbf{Supply chain favors agentic modes}: LlamaParse
Agentic and Reducto Agentic gain several points over their base configurations
on its table-heavy bills of lading and invoices. Extend Performance~v2 is the
only system that wins every domain on both metrics, and AWS~Textract is
uniformly weakest, bottoming out at 47\% per-field / 43\% per-question on
medical.

\subsection{Cost and latency}
\label{sec:results:cost}

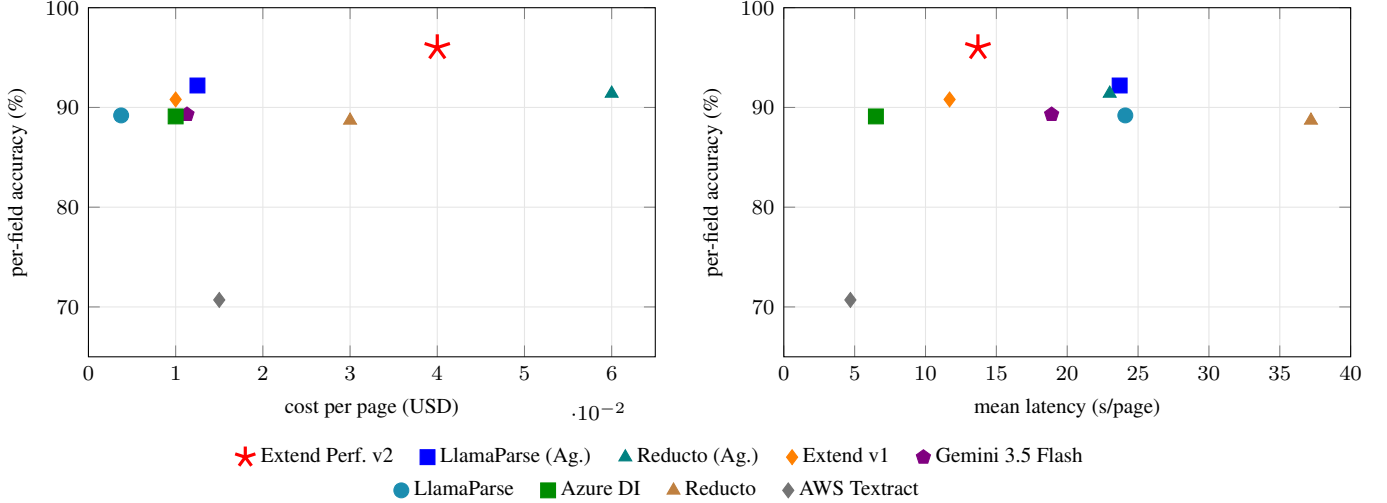
\begin{figure*}[t]
  \centering
  \begin{tikzpicture}
    \pgfplotsset{
      sysExtV2/.style ={red,            mark=star,      mark size=5pt, very thick},
      sysLPA/.style   ={blue,           mark=square*},
      sysRedA/.style  ={teal,           mark=triangle*},
      sysExtV1/.style ={orange,         mark=diamond*},
      sysGem/.style   ={violet,         mark=pentagon*},
      sysLP/.style    ={cyan!65!black,  mark=otimes*},
      sysAzure/.style ={green!55!black, mark=square*},
      sysRed/.style   ={brown,          mark=triangle*},
      sysText/.style  ={black!55,       mark=diamond*},
      sysmark/.style  ={only marks, mark size=2.8pt},
    }
    \begin{groupplot}[
      group style={group size=2 by 1, horizontal sep=1.7cm},
      width=0.52\linewidth, height=6.2cm,
      ymin=65, ymax=100,
      ylabel={per-field accuracy (\%)},
      grid=both, grid style={gray!20},
      tick label style={font=\footnotesize},
      label style={font=\footnotesize},
    ]
      \nextgroupplot[xlabel={cost per page (USD)}, xmin=0, xmax=0.065]
        \addplot[sysmark, sysLPA]   coordinates {(0.0125,92.2)};
        \addplot[sysmark, sysRedA]  coordinates {(0.060,91.4)};
        \addplot[sysmark, sysExtV1] coordinates {(0.010,90.8)};
        \addplot[sysmark, sysGem]   coordinates {(0.0113,89.3)};
        \addplot[sysmark, sysLP]    coordinates {(0.00375,89.2)};
        \addplot[sysmark, sysAzure] coordinates {(0.010,89.1)};
        \addplot[sysmark, sysRed]   coordinates {(0.030,88.7)};
        \addplot[sysmark, sysText]  coordinates {(0.015,70.7)};
        \addplot[sysmark, sysExtV2] coordinates {(0.040,96.0)};
      \nextgroupplot[xlabel={mean latency (s/page)}, xmin=0, xmax=40]
        \addplot[sysmark, sysLPA]   coordinates {(23.7,92.2)};
        \addplot[sysmark, sysRedA]  coordinates {(23.0,91.4)};
        \addplot[sysmark, sysExtV1] coordinates {(11.7,90.8)};
        \addplot[sysmark, sysGem]   coordinates {(18.9,89.3)};
        \addplot[sysmark, sysLP]    coordinates {(24.1,89.2)};
        \addplot[sysmark, sysAzure] coordinates {(6.5,89.1)};
        \addplot[sysmark, sysRed]   coordinates {(37.2,88.7)};
        \addplot[sysmark, sysText]  coordinates {(4.7,70.7)};
        \addplot[sysmark, sysExtV2] coordinates {(13.7,96.0)};
    \end{groupplot}
  \end{tikzpicture}
  \\[4pt]
  {\footnotesize\centering
    \lmarkbig{red}{star}~Extend Perf.\ v2 \quad
    \lmark{blue}{square*}~LlamaParse (Ag.) \quad
    \lmark{teal}{triangle*}~Reducto (Ag.) \quad
    \lmark{orange}{diamond*}~Extend v1 \quad
    \lmark{violet}{pentagon*}~Gemini 3.5 Flash \\[3pt]
    \lmark{cyan!65!black}{otimes*}~LlamaParse \quad
    \lmark{green!55!black}{square*}~Azure DI \quad
    \lmark{brown}{triangle*}~Reducto \quad
    \lmark{black!55}{diamond*}~AWS Textract \quad
    \par}
  \caption{Cost and latency vs.\ accuracy on the full bank, one colored marker
  per system (Extend Performance~v2 is the large red star).
  \textbf{Left:} cost vs.\ per-field accuracy
  \textbf{Right:} latency vs.\ per-field accuracy.
  Costs estimated from public numbers, actual costs may vary among the commercial providers.}
  \label{fig:pareto}
\end{figure*}

\cref{tab:cost} consolidates per-page cost, latency, and accuracy, and
\cref{fig:pareto} plots cost and latency against accuracy. Costs reflect list
prices; credit-priced systems are converted at published rates
(\cref{sec:method:cost}).\footnote{Extend Performance~v2 is priced at 2
credits/page at \$0.02/credit, i.e.\ \$0.04/page; this is the value used in the
cost-vs-accuracy frontier.}

\begin{table}[t]
  \centering
  \small
  \setlength{\tabcolsep}{4pt}
  \begin{tabular}{lrrrr}
    \toprule
    System & \$/page & s/page & P-field & P-quest. \\
    \midrule
    \textbf{Extend\ v2} & 0.040 & 13.7 & \textbf{96.0} & \textbf{90.9} \\
    LlamaParse (Agentic)      & 0.0125 & 23.7 & 92.2 & 84.5 \\
    Reducto (Agentic)         & 0.060 & 23.0 & 91.4 & 83.8 \\
    Extend v1                 & 0.010 & 11.7 & 90.8 & 82.5 \\
    Gemini 3.5 Flash          & 0.0113 & 18.9 & 89.3 & 82.2 \\
    LlamaParse                & 0.00375 & 24.1 & 89.2 & 80.8 \\
    Azure DI                  & 0.010 & 6.5 & 89.1 & 79.6 \\
    Reducto                   & 0.030 & 37.2 & 88.7 & 80.5 \\
    AWS Textract              & 0.015 & 4.7 & 70.7 & 54.0 \\
    \bottomrule
  \end{tabular}
  \caption{Cost, latency, and accuracy. Latency is the median of five
  cache-busted concurrent solo calls on a one-page document.}
  \label{tab:cost}
\end{table}

On the \textbf{cost--accuracy view}, a few systems stand out: LlamaParse
(cheapest at \$0.00375/page, 89.2\%), Extend~v1 (\$0.010, 90.8\%), LlamaParse
Agentic (\$0.0125, 92.2\%), and Extend Performance~v2 (\$0.040, 96.0\%); the
other commercial systems (Azure~DI, Reducto, Reducto Agentic, and AWS~Textract)
are each matched or beaten by an alternative that is both cheaper and more
accurate. The \textbf{latency--accuracy view} reflects a trade-off rather than a
single best system: Extend Performance~v2 anchors the high-accuracy end
(13.7\,s/page, 96.0\%), while Azure~DI offers the lowest latency among the
higher-accuracy systems (6.5\,s/page) but at meaningfully lower accuracy
(89.1\%). AWS~Textract is faster still but far less accurate, and the agentic
and single-pass LlamaParse configurations are comparatively slow.

\subsection{Open-source systems}
\label{sec:results:oss}

The eight open-source systems were run under the identical bank, extraction LLM, and
scorer, served via vLLM on an H200 GPU (except Docling, which runs in-process
on CPU). \textbf{Chandra-2 is the clear open-source leader} at 86.2\%/78.1\%,
within a few points of the weaker commercial systems and strongest on the hard
medical domain; its small backbone notably emits structured diagram
representations rather than skipping figures. \textbf{olmOCR-2 and
Nanonets-OCR-s} form a second tier in a near dead heat ($\sim$68\%
per-question), and \textbf{Docling} (a free, CPU-only library) clusters near
AWS~Textract on per-question while edging it on per-field, a strong baseline for
its cost profile. \textbf{dots.ocr} improves markedly from a single-shot
markdown configuration (65.4\%/57.2\%) to a two-stage layout-then-format recipe
(70.6\%/61.4\%). \textbf{Nemotron Nano~V2~VL}, a generalist 12B VLM, trails badly
(41.6\%/29.3\%) with characteristic OCR misreadings (e.g.\ ``\texttt{L0S5
H1ST0RY}'' for ``\texttt{LOSS HISTORY}''), confirming that a general VLM is no
substitute for an OCR-specialist model. The two PaddleOCR systems (PaddleOCR-VL
and Paddle~v3) land in the lower tier (59.6\%/48.5\% and
59.2\%/43.1\%, respectively). The open-source field as a whole still
trails the agentic commercial tier by 17+ points on per-question accuracy.

\subsection{Layout results}
\label{sec:results:layout}

\cref{tab:layout} reports layout F1 on the 1{,}500-page track, and
\cref{tab:layout-perclass} the per-class adjusted F1.

\begin{table}[t]
  \centering
  \small
  \setlength{\tabcolsep}{4pt}
  \begin{tabular}{lrrrrr}
    \toprule
    Model & Strict & Adj. & Macro & Prec. & Rec. \\
    \midrule
    \textbf{Extend v2} & \textbf{0.781} & \textbf{0.835} & 0.678 & 0.818 & 0.748 \\
    AWS Textract           & 0.643 & 0.726 & 0.478 & 0.628 & 0.657 \\
    Paddle-OCR-VL 1.5~\cite{paddleocrvl} & 0.585 & 0.680 & 0.403 & 0.661 & 0.524 \\
    Azure DI               & 0.522 & 0.637 & 0.458 & 0.431 & 0.662 \\
    dots.ocr               & 0.238 & 0.317 & 0.144 & 0.225 & 0.253 \\
    \bottomrule
  \end{tabular}
  \caption{Layout F1 on RealDocBench-Layout ($N{=}1{,}500$, nine-class
  taxonomy). ``Strict'' and ``Adj.'' are micro F1 before/after adjacency
  recovery; macro F1, precision, and recall are reported on the adjusted match.}
  \label{tab:layout}
\end{table}

\begin{table}[t]
  \centering
  \footnotesize
  \setlength{\tabcolsep}{3pt}
  \begin{tabular}{lrrrrr}
    \toprule
    Class & Extend & Textract & Paddle & Azure & dots \\
    \midrule
    text            & \textbf{0.717} & 0.532 & 0.532 & 0.381 & 0.182 \\
    heading         & \textbf{0.693} & 0.500 & 0.312 & 0.530 & 0.185 \\
    section\_heading& \textbf{0.643} & 0.426 & 0.502 & 0.480 & 0.216 \\
    header          & \textbf{0.596} & 0.274 & 0.272 & 0.386 & 0.141 \\
    footer          & \textbf{0.759} & 0.660 & 0.572 & 0.606 & 0.102 \\
    page\_number    & \textbf{0.824} & 0.753 & 0.652 & 0.764 & 0.000 \\
    figure          & \textbf{0.784} & 0.555 & 0.329 & 0.560 & 0.183 \\
    table           & \textbf{0.761} & 0.672 & 0.611 & 0.588 & 0.374 \\
    key\_value      & \textbf{0.568} & 0.121 & 0.000 & 0.000 & 0.000 \\
    \bottomrule
  \end{tabular}
  \caption{Per-class adjusted F1 (nine-class taxonomy). Three systems score
  \texttt{key\_value}${=}0$ because their public output vocabulary has no
  equivalent class; dots.ocr also scores \texttt{page\_number}${=}0$ for the
  same structural reason.}
  \label{tab:layout-perclass}
\end{table}

Extend~v2.0.0 leads on both strict (0.781) and adjusted (0.835) F1 and on every
per-class score. Adjacency recovery lifts all systems, typically by 5--10
points of micro F1, confirming that a non-trivial share of apparent errors are
boundary disagreements rather than missed content. The \texttt{key\_value} row
is the most discriminating: three of five systems score zero because their
public taxonomy has no key--value class at all, illustrating both the value of
including that class and the symmetric-fairness design that scores every system
on the full taxonomy (\cref{sec:method:layout}) without double-penalizing a
structurally absent class.

\section{Discussion}
\label{sec:discussion}

\paragraph{Findings.}
Two findings stand out from \cref{sec:results}. First, \textbf{difficulty
is concentrated, not uniform}: medical documents and finance forms are where
systems diverge, while clean mortgage forms are nearly saturated. A benchmark
that averages over these strata hides exactly the behavior practitioners care
about. Second, \textbf{accuracy must be read against
cost and latency}: the cost-accuracy-latency analysis shows several distinct operating points
rather than one dominant system, and an ``agentic'' configuration that wins on
easy content can still erode on the hard tail.

\paragraph{Open-source trajectory.}
The open-source gap is closing at the top: Chandra-2 now sits within a few
points of the weaker commercial systems and leads on the hardest domain. But
the field as a whole still trails the agentic commercial tier by a wide margin
on the strict per-question metric, and generalist VLMs remain unreliable as
parsers. However, we expect this frontier to move quickly. As an example, the recently released Opus 4.8
outperforms LlamaParse but is more expensive. We will continue to refresh the
open-source leaderboard as new systems appear.

\paragraph{Limitations.}
\rdb has several limitations we report transparently. (i) The shared extraction LLM
is a vision-language model; although it is identical across all systems and the
scorer is deterministic, it can occasionally misread a parser's markdown. Gold
answers, by contrast, were drafted by two independent models (Sonnet and
Gemini), difficulty-graded, and manually verified; rather than
discarding every model disagreement we deliberately re-included a portion so the bank is not skewed
toward only cleanly answerable fields. Because we re-included only a sample of the disagreements, the hardest cases are represented but not exhaustively. (ii) Latency is measured on a small set of single one-page
documents under cold-cache conditions and may not reflect sustained,
multi-page throughput. (iii) Costs are list prices and ignore enterprise
discounts. (iv) The domain distribution is intentionally uneven and reflects
our sourcing priorities rather than a census of all document types. (v) To
respect privacy, a portion of the medical/healthcare and tax documents are real,
blank form templates populated with synthetic persona data
(\cref{sec:construction:sourcing}) rather than real filled instances (tax also
includes real documents from historical public archives); for the synthesized
items the layouts are authentic but the field \emph{contents} are fabricated,
which may not capture every irregularity of genuine completed records
(idiosyncratic handwriting, real-world data entry errors). (vi) As the
benchmark is authored by a vendor whose systems perform well, we mitigate
conflict-of-interest concerns by holding every system to an identical
extraction-and-scoring protocol, publishing the taxonomy mappings and adapters,
and releasing the data and harness for independent reproduction. (vii) To keep
the comparison uniform, every parser is invoked through a single fixed prompt
and configuration rather than one tuned per system. Several systems, especially
the general-purpose VLMs are known to be
sensitive to prompting, so their reported numbers should be read as the
performance of a reasonable default rather than an upper bound: targeted prompt
optimization or system-specific configuration would likely raise some of these
scores. We release the prompts and adapters
so that others can explore tuned variants.

\paragraph{Conclusion.}
\rdb evaluates document parsing on the task that matters in regulated
workflows, retrieving correct field values from real, messy documents, and
pairs it with a layout track on a public taxonomy. Across eighteen systems it
exposes a structure that single-number benchmarks miss: large gaps between
tiers, concentrated difficulty by domain and capability, and explicit
cost/latency trade-offs. We release the datasets, parser
adapters, and evaluation harness, and invite the community to extend the
benchmark with new documents, domains, and systems.

{
    \small
    \bibliographystyle{ieeenat_fullname}
    \bibliography{main}
}

\appendix
\onecolumn
\section{Privacy-Preserving Document Synthesis}
\label{app:synthesis}

This appendix details the pipeline used to construct the synthesized portion of
the medical/healthcare and tax tracks, summarized in
\cref{sec:construction:sourcing}.

\paragraph{Motivation and sourcing.}
The medical/healthcare and tax genres can carry personally identifiable and
protected health information. Where redistributable real instances are
available, for example tax documents drawn from historical public
archives, we use them directly. Where they are not, we avoid real filled
records and instead start from \emph{real, blank form templates}, preserving the
authentic layout, fields, and visual structure of the genre, and populate them
with synthetic content. This is distinct from the low-fidelity synthetic pages
rejected during integration (\cref{sec:construction:sourcing}): the layouts are
real and the rendered fill is genre-faithful; only the underlying
\emph{information} is fabricated, by design, to protect privacy.

\paragraph{Per-template pipeline.}
For each blank template we run a fixed sequence:
\begin{enumerate}
  \item \textbf{Schema extraction.} A vision model reads each page and emits a
    typed field schema for the form.
  \item \textbf{Template analysis.} A second pass characterizes how the form is
    meant to be completed (handwriting convention, checkbox density, and
    signature locations) to inform realistic persona design.
  \item \textbf{Persona generation.} We synthesize a set of diverse fictitious
    personas (multiple per template), each with internally consistent attributes
    (names, dates, identifiers, clinical or financial values) and explicit
    checkbox/handwriting profiles. Personas use clearly synthetic data (e.g.\
    \texttt{555} phone exchanges, placeholder identifiers) and no real
    individuals.
  \item \textbf{Filling.} Templates are filled two ways: (i) programmatically
    via the form's \texttt{AcroForm} fields for typed/digital input, and
    (ii) with an image-editing model (\texttt{gpt-image-2}) to render
    handwritten and stamped-style entries on forms that lack fillable fields.
  \item \textbf{Fidelity screening.} Every synthesized page is scored by a
    vision-language rubric that compares it against the \emph{original blank
    template} along four axes: \emph{form preservation} (are the printed labels,
    borders, and rules intact?), \emph{value plausibility}, \emph{checkbox
    discipline} (only intended boxes checked), and \emph{handwriting realism}.
    Pages on which the image-editing model has altered the template rather than
    merely filling it are caught and refilled.
\end{enumerate}

\paragraph{Consequences for gold and privacy.}
Because the persona attributes are the values written onto the form, they serve
as ground-truth answers \emph{by construction}: gold for these documents is
known a priori and requires substantially less manual verification than the
model-drafted gold used elsewhere (\cref{sec:construction:qa}). On privacy, we
take care not to introduce real PII into the synthesized documents, and the
blank form templates themselves are real but standardized, publicly available
forms rather than any individual's completed record, so reusing them does not
expose personal information.

\section{Example QA Items by Domain}
\label{app:examples}

To complement the single item in \cref{fig:example-qa}, we show one real
question from each of the four domains alongside the source page and the actual
output of the strongest system (Extend Performance~v2). Every item is drawn
verbatim from the released bank: the gold answer is the typed
\texttt{gold\_dict}, and the Extend~v2 column is its cached prediction under the
shared extraction LLM and scorer. Three of the four are answered correctly and
one (medical) is missed, illustrating both the structural difficulty of the
questions and the type-tolerant matching that, for example, scores the string
\texttt{"unselected"} equal to the boolean \texttt{false}. Each
per-field comparison is marked \cmark{} (correct) or \xmark{} (incorrect).

\begin{figure*}[tp]
  \centering
  \begin{minipage}[t]{0.30\linewidth}
    \centering\vspace{0pt}
    \fbox{\includegraphics[width=0.95\linewidth]{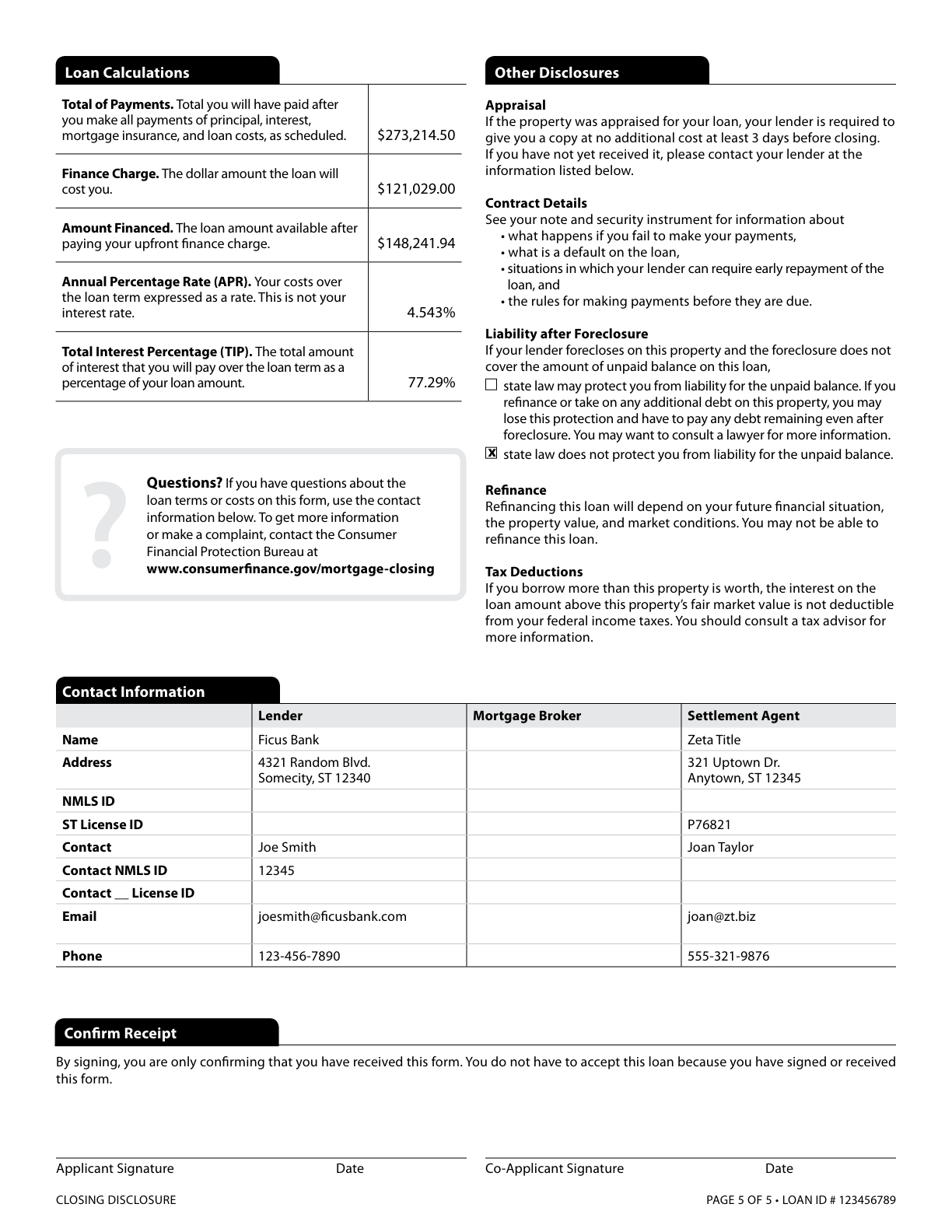}}
  \end{minipage}\hfill
  \begin{minipage}[t]{0.66\linewidth}
    \vspace{0pt}\small
    \textbf{Domain:} mortgage \quad\textbf{Doc:} CFPB Closing Disclosure
    (H-25E, refinance)\\[3pt]
    \textbf{Question.} \textit{From the Contact Information table, please
    provide the Lender's email address and physical address, as well as the ST
    License ID for the Settlement Agent.}\\[4pt]
    \footnotesize
    \begin{tabular}{@{}p{2.6cm}p{3.3cm}p{3.3cm}c@{}}
      \toprule
      Field & Gold & Extend v2 & \\
      \midrule
      \ttfamily lender\_email & joesmith@ficusbank.com & joesmith@ficusbank.com & \cmark \\
      \ttfamily lender\_address & 4321 Random Blvd. Somecity, ST 12340 & 4321 Random Blvd. Somecity, ST 12340 & \cmark \\
      \ttfamily settlement\_agent\_st\_license\_id & P76821 & P76821 & \cmark \\
      \bottomrule
    \end{tabular}\\[4pt]
    \textbf{Capabilities:} {\small field\_value\_pairing, multi\_column\_grid,
    row\_binding, form\_region}
  \end{minipage}
  \caption{\textbf{Mortgage example.} A multi-field question that binds three
  values from a contact-information grid; the parser must keep each label
  aligned with its value across columns.}
  \label{fig:ex-mortgage}
\end{figure*}

\begin{figure*}[tp]
  \centering
  \begin{minipage}[t]{0.30\linewidth}
    \centering\vspace{0pt}
    \fbox{\includegraphics[width=0.95\linewidth]{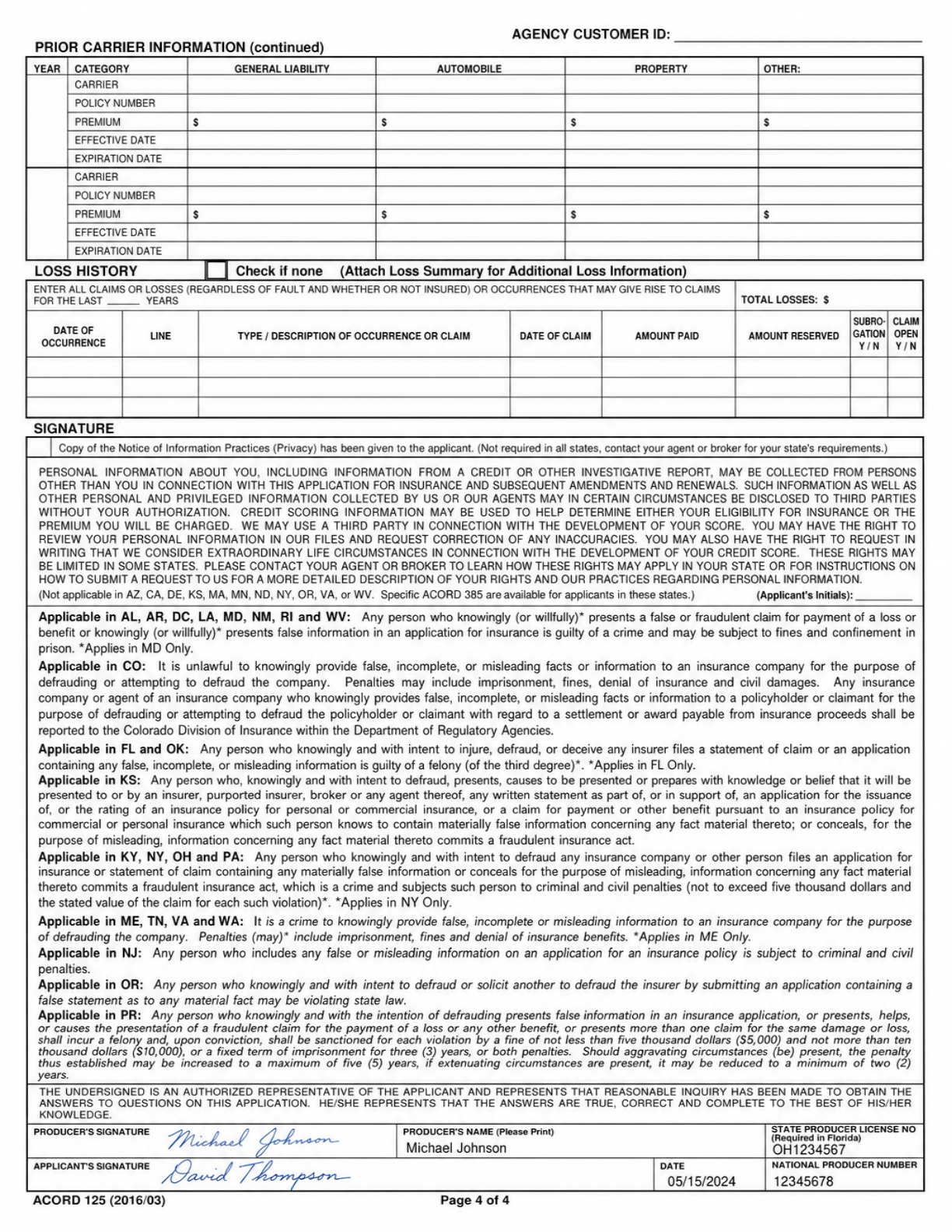}}
  \end{minipage}\hfill
  \begin{minipage}[t]{0.66\linewidth}
    \vspace{0pt}\small
    \textbf{Domain:} finance \quad\textbf{Doc:} ACORD 125/126/140 (commercial
    insurance application)\\[3pt]
    \textbf{Question.} \textit{In the LOSS HISTORY section, is the `Check if
    none' box selected, and in the table, what is the header text for the
    column between DATE OF CLAIM and AMOUNT RESERVED?}\\[4pt]
    \footnotesize
    \begin{tabular}{@{}p{3.5cm}p{2.2cm}p{2.5cm}c@{}}
      \toprule
      Field & Gold & Extend v2 & \\
      \midrule
      \ttfamily check\_if\_none & false & "unselected" & \cmark \\
      \ttfamily column\_header\_between\_date\_and\_reserved & AMOUNT PAID & AMOUNT PAID & \cmark \\
      \bottomrule
    \end{tabular}\\[2pt]
    {\footnotesize The string \texttt{"unselected"} is matched to boolean
    \texttt{false} by the type-tolerant scorer.}\\[4pt]
    \textbf{Capabilities:} {\small checkbox\_state, handdrawn\_check,
    multi\_column\_grid, row\_binding, table\_structure}
  \end{minipage}
  \caption{\textbf{Finance example.} A checkbox state plus a table column
  header read from the same region; the printed ``Check if none'' box is a
  common decoy.}
  \label{fig:ex-finance}
\end{figure*}

\begin{figure*}[tp]
  \centering
  \begin{minipage}[t]{0.30\linewidth}
    \centering\vspace{0pt}
    \fbox{\includegraphics[width=0.95\linewidth]{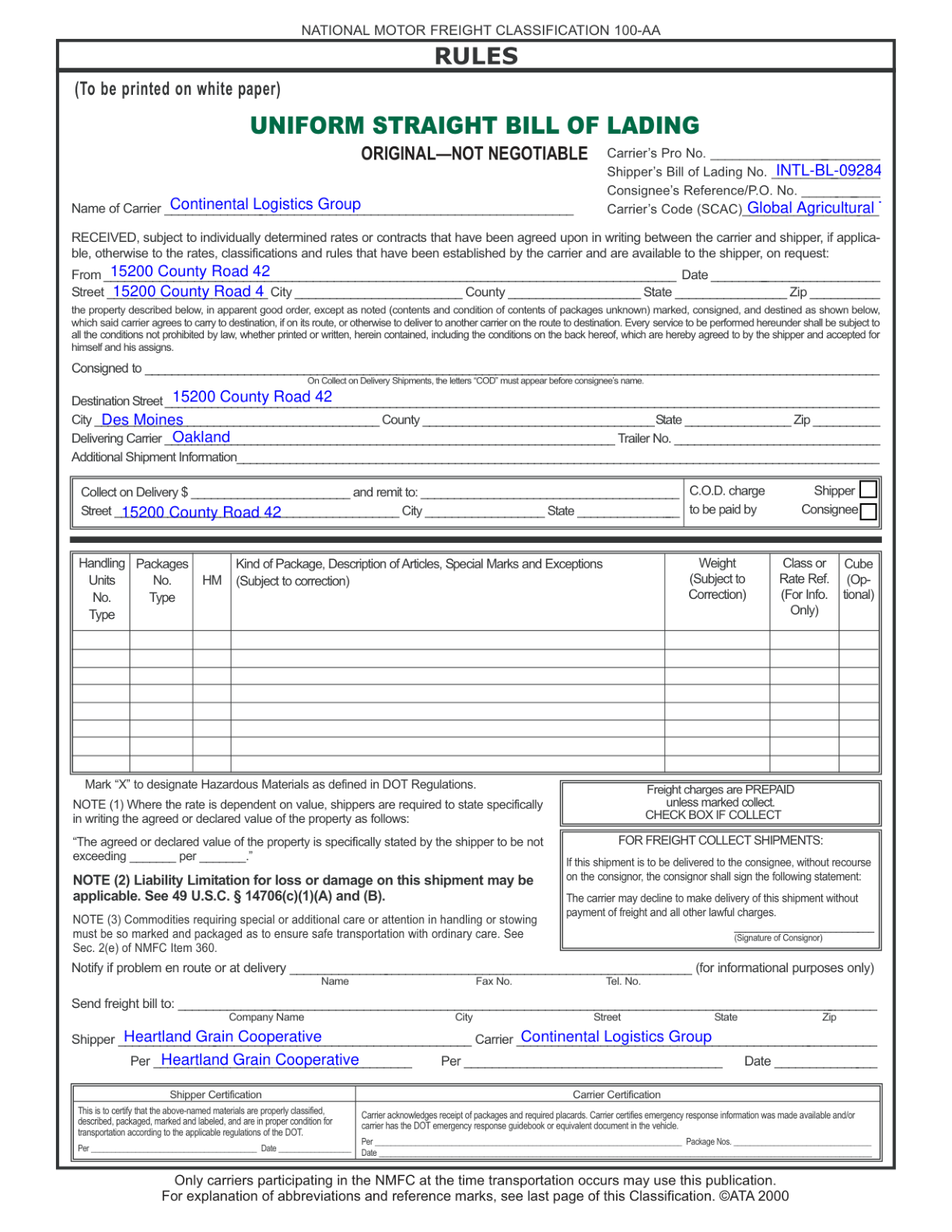}}
  \end{minipage}\hfill
  \begin{minipage}[t]{0.66\linewidth}
    \vspace{0pt}\small
    \textbf{Domain:} supply\_chain \quad\textbf{Doc:} ARCB Uniform Straight
    Bill of Lading\\[3pt]
    \textbf{Question.} \textit{In the `C.O.D. charge to be paid by' section,
    what are the checkbox states for Shipper and Consignee?}\\[4pt]
    \footnotesize
    \begin{tabular}{@{}p{2.8cm}p{2.0cm}p{2.6cm}c@{}}
      \toprule
      Field & Gold & Extend v2 & \\
      \midrule
      \ttfamily shipper\_checkbox & false & "unchecked" & \cmark \\
      \ttfamily consignee\_checkbox & false & "unchecked" & \cmark \\
      \bottomrule
    \end{tabular}\\[4pt]
    \textbf{Capabilities:} {\small checkbox\_state, field\_value\_pairing,
    parallel\_columns}
  \end{minipage}
  \caption{\textbf{Supply-chain example.} Two parallel checkbox columns
  (Shipper vs.\ Consignee) that must be read independently; both are unchecked.}
  \label{fig:ex-supply}
\end{figure*}

\begin{figure*}[tp]
  \centering
  \begin{minipage}[t]{0.30\linewidth}
    \centering\vspace{0pt}
    \fbox{\includegraphics[width=0.95\linewidth]{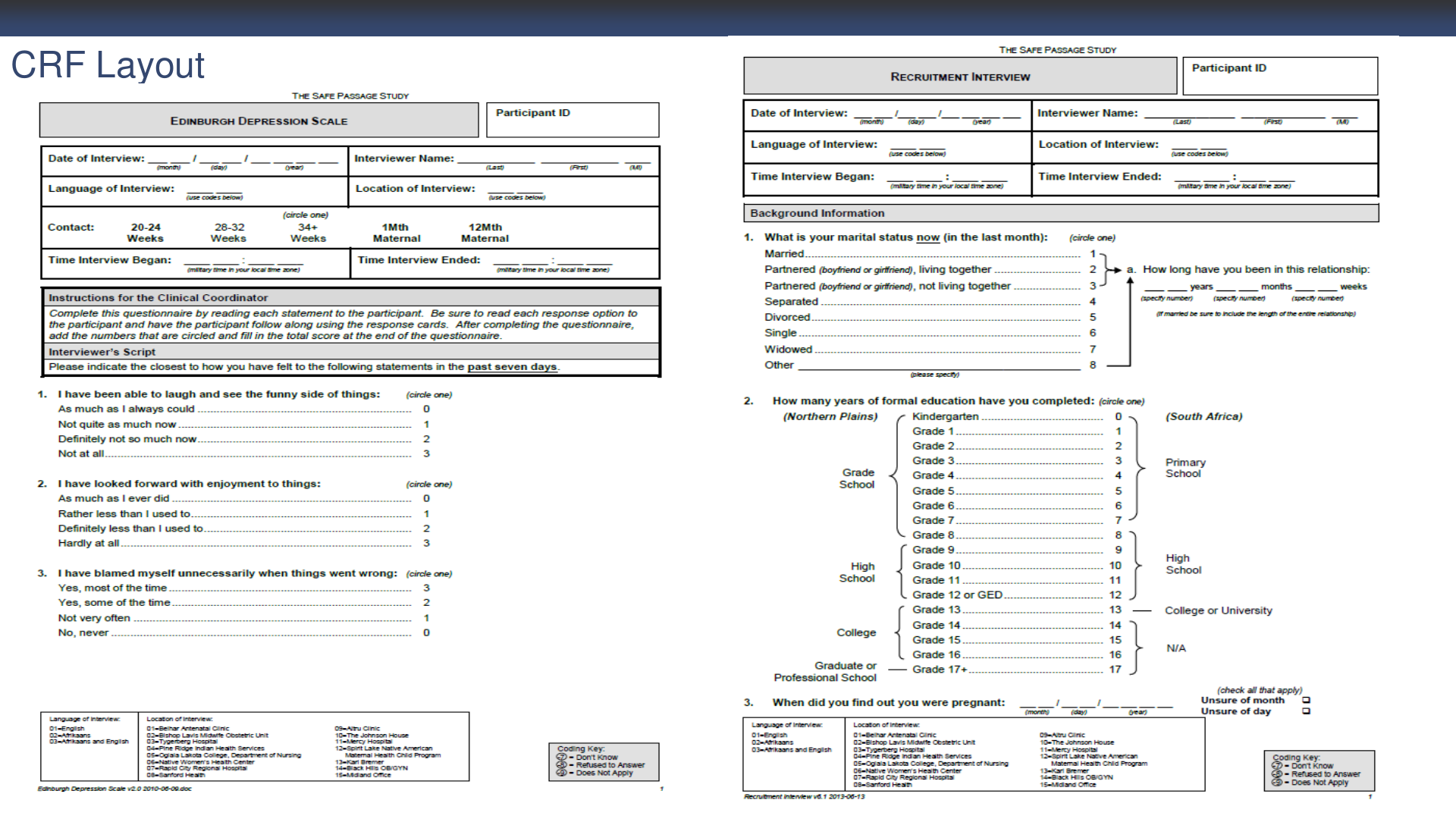}}
  \end{minipage}\hfill
  \begin{minipage}[t]{0.66\linewidth}
    \vspace{0pt}\small
    \textbf{Domain:} medical\_healthcare \quad\textbf{Doc:} BU BEDAC clinical
    Case Report Form\\[3pt]
    \textbf{Question.} \textit{In the `Coding Key' box on the Recruitment
    Interview page, what are the numeric codes assigned to `Don't Know' and
    `Refused to Answer'?}\\[4pt]
    \footnotesize
    \begin{tabular}{@{}p{3.0cm}p{2.0cm}p{2.4cm}c@{}}
      \toprule
      Field & Gold & Extend v2 & \\
      \midrule
      \ttfamily dont\_know\_code & 77 & null & \xmark \\
      \ttfamily refused\_to\_answer\_code & 98 & 65 & \xmark \\
      \bottomrule
    \end{tabular}\\[2pt]
    {\footnotesize A genuine miss: the small printed coding key is read wrong,
    the failure mode the strict per-question metric is designed to surface.}\\[4pt]
    \textbf{Capabilities:} {\small circled\_choice, field\_value\_pairing,
    form\_region}
  \end{minipage}
  \caption{\textbf{Medical example.} Two numeric codes read from a printed
  ``Coding Key'' box. Even the strongest system misses both, a reminder that
  dense small-print regions remain hard.}
  \label{fig:ex-medical}
\end{figure*}

\section{Confidence Intervals}
\label{app:ci}

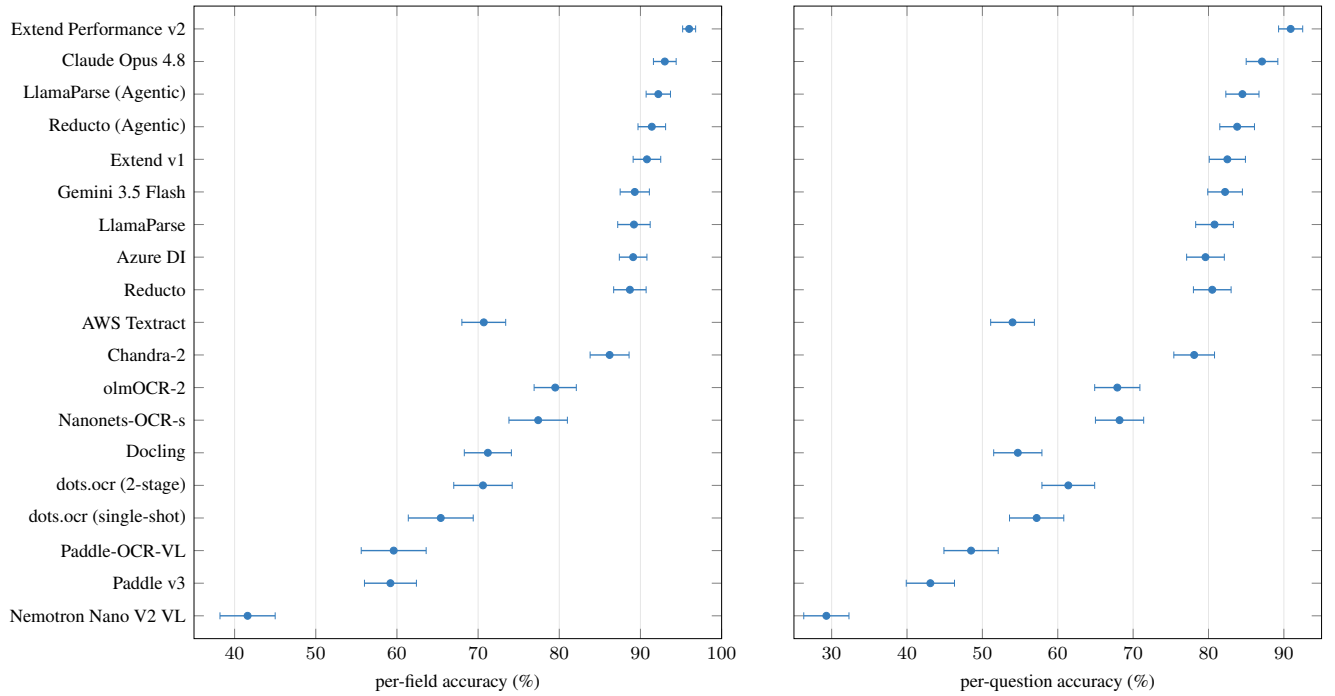
\begin{figure*}[t]
  \centering
  \resizebox{\textwidth}{!}{%
  \begin{tikzpicture}
    \begin{groupplot}[
      group style={group size=2 by 1, horizontal sep=1.1cm},
      width=0.55\linewidth, height=11.2cm,
      ymin=0.3, ymax=19.7,
      ytick={1,2,3,4,5,6,7,8,9,10,11,12,13,14,15,16,17,18,19},
      xmajorgrids, grid style={gray!20},
      tick label style={font=\footnotesize},
      label style={font=\footnotesize},
      every axis plot/.append style={
        only marks, mark=*, mark size=1.5pt, cvprblue,
        error bars/x dir=both, error bars/x explicit,
        error bars/error bar style={cvprblue, line width=0.6pt}},
    ]
      \nextgroupplot[xlabel={per-field accuracy (\%)}, xmin=35, xmax=100,
        yticklabels={Nemotron Nano V2 VL, Paddle v3, Paddle-OCR-VL,
          dots.ocr (single-shot), dots.ocr (2-stage), Docling, Nanonets-OCR-s,
          olmOCR-2, Chandra-2, AWS Textract, Reducto, Azure DI, LlamaParse,
          Gemini 3.5 Flash, Extend v1, Reducto (Agentic), LlamaParse (Agentic),
          Claude Opus 4.8, Extend Performance v2}]
        \addplot[] coordinates {
          (96.0,19) +- (0.80,0)  (93.0,18) +- (1.40,0)  (92.2,17) +- (1.50,0)
          (91.4,16) +- (1.70,0)  (90.8,15) +- (1.70,0)  (89.3,14) +- (1.80,0)
          (89.2,13) +- (2.00,0)  (89.1,12) +- (1.70,0)  (88.7,11) +- (2.00,0)
          (70.7,10) +- (2.70,0)  (86.2,9)  +- (2.40,0)  (79.5,8)  +- (2.60,0)
          (77.4,7)  +- (3.60,0)  (71.2,6)  +- (2.90,0)  (70.6,5)  +- (3.60,0)
          (65.4,4)  +- (4.00,0)  (59.6,3)  +- (4.00,0)  (59.2,2)  +- (3.20,0)
          (41.6,1)  +- (3.40,0) };
      \nextgroupplot[xlabel={per-question accuracy (\%)}, xmin=25, xmax=95,
        yticklabel=\empty]
        \addplot[] coordinates {
          (90.9,19) +- (1.60,0)  (87.1,18) +- (2.10,0)  (84.5,17) +- (2.20,0)
          (83.8,16) +- (2.30,0)  (82.5,15) +- (2.40,0)  (82.2,14) +- (2.30,0)
          (80.8,13) +- (2.50,0)  (79.6,12) +- (2.50,0)  (80.5,11) +- (2.50,0)
          (54.0,10) +- (2.90,0)  (78.1,9)  +- (2.70,0)  (67.9,8)  +- (3.00,0)
          (68.2,7)  +- (3.20,0)  (54.7,6)  +- (3.20,0)  (61.4,5)  +- (3.50,0)
          (57.2,4)  +- (3.60,0)  (48.5,3)  +- (3.60,0)  (43.1,2)  +- (3.20,0)
          (29.3,1)  +- (3.00,0) };
    \end{groupplot}
  \end{tikzpicture}}
  \caption{\textbf{95\% document-clustered bootstrap confidence intervals}
  ($B{=}10{,}000$) for every row of the QA leaderboard.
  \textbf{Left:} per-field accuracy; \textbf{Right:} per-question
  accuracy. Markers are point estimates; whiskers are the $2.5$/$97.5$
  percentile interval. Overlapping intervals indicate differences that are not
  statistically resolved on the current bank, e.g.\ the single-pass cluster of
  Gemini, LlamaParse, Azure~DI, and Reducto; Extend Performance~v2's per-field
  interval is disjoint from all others.}
  \label{fig:forest}
\end{figure*}

We report 95\% confidence intervals for every QA system using the
document-clustered bootstrap of \cref{sec:method:qa} (resample the 581
documents with replacement, $B{=}10{,}000$, percentile interval).
\cref{fig:forest} plots the per-field and per-question intervals for every row
of the QA leaderboard (\cref{tab:leaderboard}).

\end{document}